\documentclass[11pt,letterpaper]{mystyle}

\usepackage{algorithmic}
\usepackage{algorithm}
\usepackage{array}
\usepackage{stfloats}
\usepackage{verbatim}
\usepackage[numbers]{natbib}
\usepackage{nicefrac}       
\usepackage{multirow}
\usepackage{tablefootnote}
\usepackage{makecell}
\usepackage{xparse}
\usepackage{fontawesome5}
\usepackage{bxcoloremoji}
\usepackage{twemojis}
\usepackage{bbding}
\usepackage{lipsum}

\usepackage{forest}
\useforestlibrary{edges}
\usepackage{tikz}

\definecolor{deepgreen}{HTML}{057311}
\definecolor{AgentIndigo}{HTML}{3F51B5}
\definecolor{AgentIndigoLight}{HTML}{E8EAF6}
\definecolor{AgentAmber}{HTML}{FF8F00}
\definecolor{AgentAmberLight}{HTML}{FFF3E0}

\usepackage[table]{xcolor}
\definecolor{RowGray}{HTML}{F7F7F7}

\tcbset{
  agentcontrib/.style={
    colback=MetaBoxBg,
    colframe=MetaBoxBg,
    colbacktitle=MetaBoxFg!8!MetaBoxBg,
    coltext=MetaBoxFg,
    coltitle=MetaBoxFg,
    boxrule=0.9pt, arc=2mm,
    left=3mm, right=3mm, top=2mm, bottom=2mm,
    fonttitle=\sffamily\fontseries{bx}\selectfont,
    title=Contributions
  }
}

\runningtitle{AI Smart Glasses for Wearable Intelligence}

\title{%
\textbf{AI Smart Glasses for Wearable Intelligence: From Egocentric Sensing to Agentic Personalization}
}

\author{
   \normalfont
    Xu Yuan$^{1 \dag}$ \quad
    Yi Wang$^{1 \dag}$ \quad
    Zhuohang Jiang$^{1}$ \quad
    Haohao Qu$^{1}$ \quad
    Yujuan Ding$^{1}$ \quad
    Shanru Lin$^{1}$ \quad \\
    Guoliang Xing$^{2}$ \quad 
    Hongxia Yang$^{1}$ \quad
    Jiannong Cao$^{1}$ \quad
    Qing Li$^{1}$ \quad
    Wenqi Fan$^{1}$$^{\coloremojicode{2709}}$ \quad
   \\
   \vspace{12pt}
   \small
   \raisebox{-0.25ex}{\includegraphics[height=2ex]{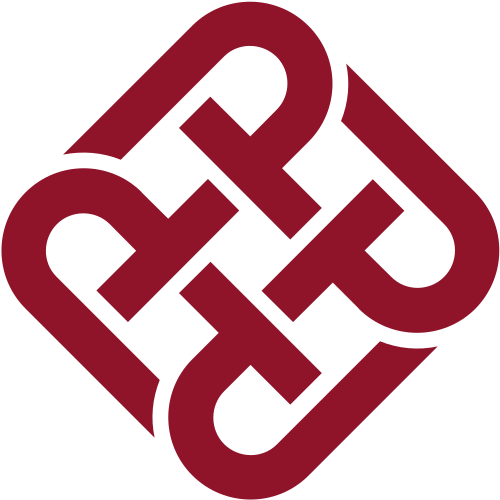}}$^{1}$The Hong Kong Polytechnic University ~~
   \raisebox{-0.25ex}{\includegraphics[height=2ex]{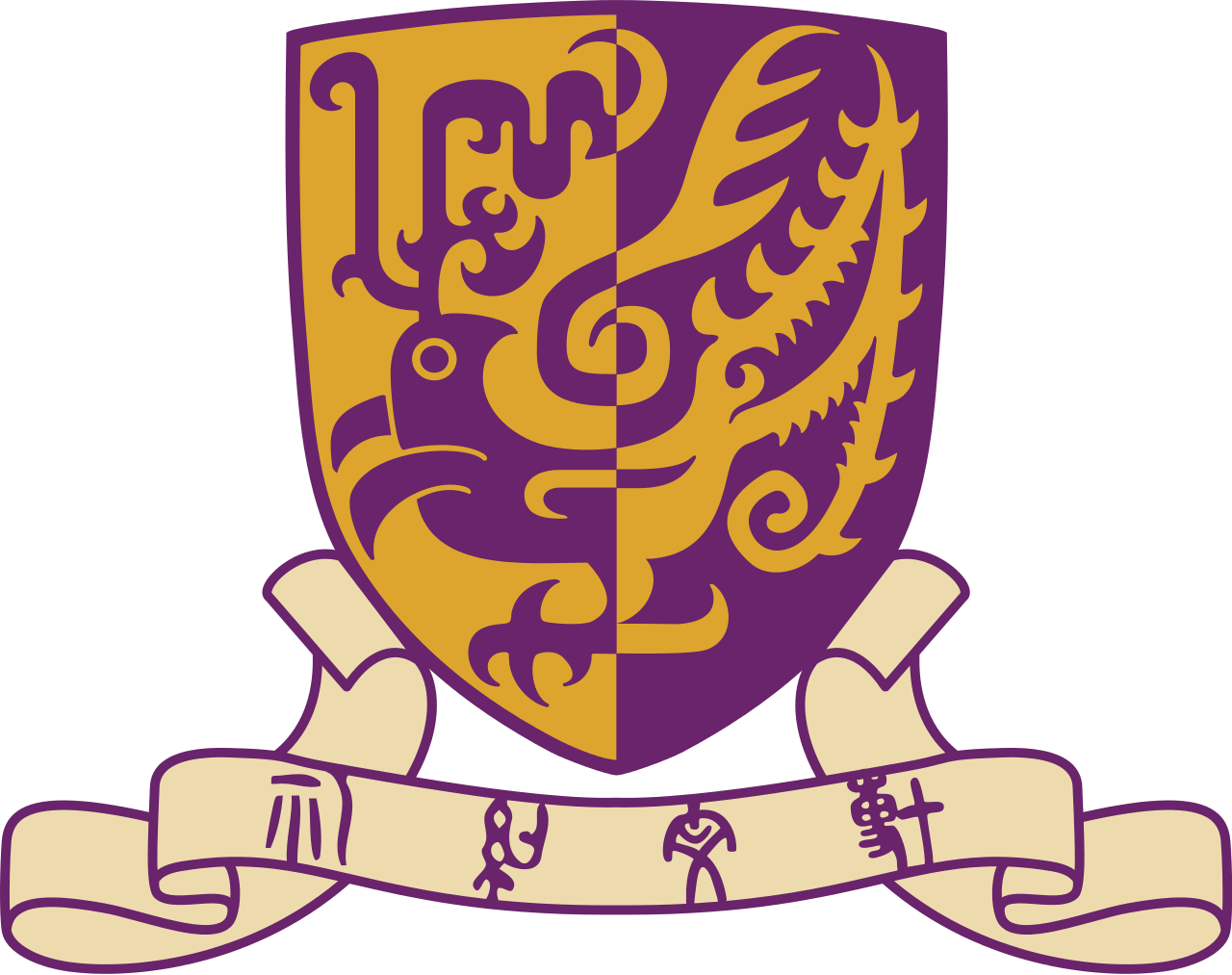}}$^{2}$The Chinese University of Hong Kong
   \\
   $^{\dag}$ \textit{Core Contributors}, \quad
   $^{\coloremojicode{2709}}$ \textit{Correspondence: wenqifan03@gmail.com}
}

\date{\vspace{-3ex}}

\begin{document}

\begin{abstract}
\textbf{\large Abstract:} 
Recent advances in artificial intelligence (AI) are reshaping smart glasses from egocentric capture and display devices into platforms for wearable intelligence. Smart glasses increasingly serve as wearable AI systems that connect first-person observation with real-time assistance under strict form-factor constraints. We frame this transition through the lens of \emph{AI smart glasses} and define them as a system-level concept in which egocentric sensing, resource-aware computing, intelligent reasoning, multimodal interaction, and real-world application constraints are co-designed for personalized assistance in the physical world. To systematically study this perspective, we organize the survey around four connected dimensions. First, we examine the hardware foundation that bounds sensing, computation, feedback delivery, and sustained deployment. Second, we study wearable intelligence, where egocentric signals are transformed into perceptual, contextual, and agentic capabilities. Third, we discuss interaction design, through which users request, receive, correct, and regulate assistance during ongoing activity. Fourth, we analyze application scenarios across healthcare, accessibility, situated learning, daily life assistance, cultural tourism, and industrial support, showing how domain requirements reshape system design and evaluation. We further identify five cross-cutting research challenges for future AI smart glasses: next-generation hardware, trustworthy egocentric intelligence, lifelong personalized memory, proactive intelligence, and embodied foundation models. By centering smart glasses as wearable-intelligence platforms, this survey provides a unified framework for organizing technologies, applications, and open challenges in this emerging area.

\vspace{5mm}
{\faBullseye}~\textbf{Keywords}: AI Smart Glasses, Egocentric Sensing, Multimodal Agents, Wearable Intelligence, Wearable Interaction, Personalized Intelligent Assistant\\
\faGithub~\textbf{Github}: \url{https://github.com/xandery-geek/Awesome-AI-Smart-Glasses-Papers}
\end{abstract}

\maketitle

\vspace{-3mm}
\section{Introduction}

Digital computing increasingly supports information access, interpersonal communication, work coordination, and skill acquisition. However, much of this support still depends on devices that must be held, watched, or explicitly operated, interrupting physical activity and limiting access to the user's immediate context. Wearable technologies address this limitation by placing sensing, computation, and feedback on the body, enabling digital assistance that is more continuous, contextual, and compatible with mobile or hands-busy situations~\cite{starner2002wearable,patel2012review}. Among wearable devices, smart glasses are distinctive because their sensors and feedback channels are aligned with the user's visual field and auditory space. This egocentric form factor allows them to observe the world from the wearer's perspective and provide hands-free assistance without requiring a full shift of attention to a separate device~\cite{grauman2022ego4d,lee2018interaction}.

\begin{figure}[t]
    \centering
    \includegraphics[width=0.95\textwidth]{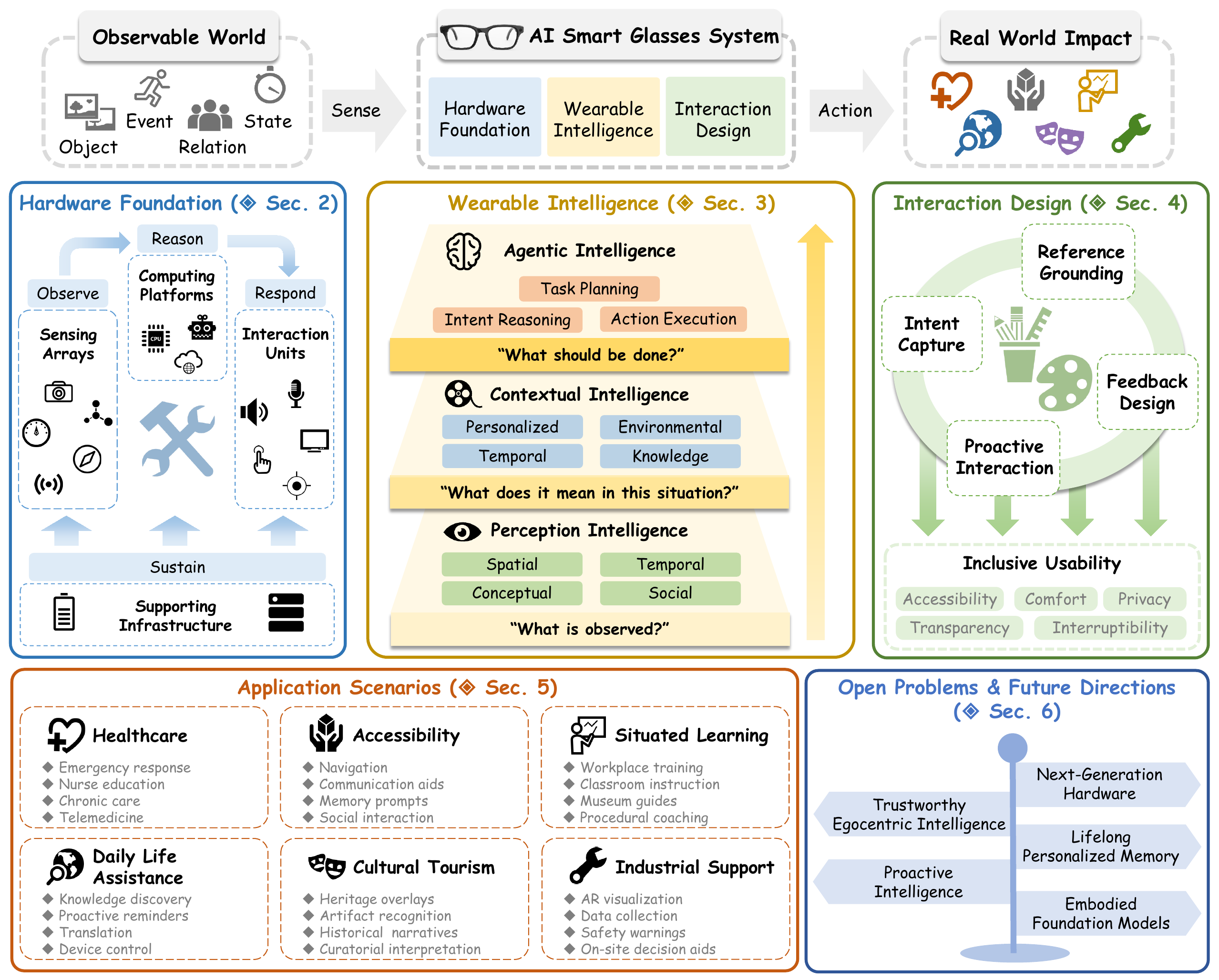}
    \caption{Overview of this survey. AI smart glasses are organized as integrated wearable-intelligence platforms that connect hardware foundation, wearable intelligence, interaction design, application scenarios, and cross-cutting research challenges.}
    \label{fig:overview}
\end{figure}

Recent advances in artificial intelligence (AI), including large multimodal models~\cite{liu2023visual,yuan2024osprey,yuan2025instruction}, retrieval-augmented generation~\cite{fan2024survey,wang2026mixture}, and AI agents~\cite{ning2025survey,jiang2026abagent,wu2026datamart}, are changing the role of smart glasses in wearable computing. Rather than serving mainly as wearable sensing and display devices, smart glasses are becoming platforms for wearable intelligence that combine the glasses form factor with AI-driven perception, reasoning, and action. Work such as EgoLife~\cite{yang2025egolife}, SUPERGLASSES~\cite{jiang2026superglasses}, and VisionClaw~\cite{liu2026visionclaw} illustrates this emerging direction across daily memory, question answering, and agentic service scenarios. In these examples, smart glasses convert egocentric observations into task-relevant states, link those states to external or personal knowledge, infer what the wearer may need, and deliver assistance through external tools, connected services, or situated feedback. In this survey, we use \emph{AI smart glasses} to denote this emerging class and define it as a system-level concept in which egocentric sensing, resource-aware computing, contextual and agentic intelligence, multimodal interaction, and real-world application constraints are co-designed to support personalized intelligent assistance in the physical world.

Realizing this vision requires coordination across device capabilities, model inference, user interaction, and deployment constraints. Each dimension is shaped by the wearable form factor. Richer sensing may improve perceptual coverage but increases power consumption, device complexity, and privacy concerns for wearers and bystanders; larger models and cloud-enhanced processing may improve contextual and agentic reasoning but add latency, energy cost, and connectivity dependence; and more proactive or expressive assistance may reduce user effort but can also interrupt attention, expose sensitive context, or act on uncertain intent. Taken together, these trade-offs show that AI smart glasses must be designed as an integrated system: they need to transform egocentric sensing into contextual interpretation and situated support while remaining controllable, deployable, and usable in daily wearable contexts.

To systematically characterize AI smart glasses, this survey focuses on smart glasses and related egocentric wearable devices that contribute to the sense--understand--act process that supports situated assistance under wearable constraints. As illustrated in Figure~\ref{fig:overview}, we organize the main body into four connected analytical parts and a cross-cutting research outlook. This organization follows the path by which wearable intelligence becomes operational in smart glasses: a physical and computational substrate supports perceptual, contextual, and agentic capabilities; interaction design exposes those capabilities to wearers; application scenarios examine how they are shaped by domain-specific requirements; and future directions synthesize the unresolved research agenda across these components.

First, the \emph{Hardware Foundation} (Section~\ref{sec:hardware-foundation}) studies the physical and computational substrate on which AI smart glasses operate. Sensing arrays define what the device can observe from the wearer's perspective, and computing platforms determine how inference is coordinated across local and cloud resources. Interaction units shape how assistance is delivered and controlled, while supporting infrastructure sustains the device during daily use. This part establishes a basic design principle: the intelligence of smart glasses is bounded by the physical constraints of wearable deployment.

Building on this substrate, \emph{Wearable Intelligence} (Section~\ref{sec:wearable-intelligence}) examines how AI smart glasses convert egocentric signals into assistance. We organize this process into three levels: \emph{Perception Intelligence} extracts spatial, temporal, conceptual, and social structure from egocentric observations; \emph{Contextual Intelligence} relates these observations to personalized, environmental, temporal, and knowledge-grounded context; \emph{Agentic Intelligence} moves from understanding to action through intent reasoning, planning, and execution. These levels form a closed loop from observation to response, rather than a collection of isolated recognition models.

Next, \emph{Interaction Design} (Section~\ref{sec:interaction-design}) examines how wearers request, receive, and correct assistance via AI smart glasses during ongoing activities. Since the interaction between wearers and smart glasses is mobile, situated, and attention-constrained, we organize it around intent capture, reference grounding, feedback design, proactive interaction, and inclusive usability. Together, these aspects show how model capability is exposed through interfaces that fit attention, activity, ability, and social context.

Fourth, \emph{Application Scenarios} (Section~\ref{sec:applications}) examine how the framework is instantiated across healthcare, accessibility, situated learning, daily-life assistance, cultural tourism, and industrial support, where assistance is shaped by distinct domain-specific user needs. Finally, \emph{Open Problems and Future Directions} (Section~\ref{sec:future-direction}) outlines five cross-cutting challenges for building AI smart glasses as personalized, proactive, and embodied assistants: next-generation hardware, trustworthy egocentric intelligence, lifelong personalized memory, proactive intelligence, and embodied foundation models.

This organization distinguishes our survey from adjacent reviews, which provide important foundations but cover only part of the system-level perspective on AI smart glasses. Egocentric-vision surveys organize egocentric datasets, perception tasks, and recognition challenges~\cite{li2026challenges}; smart-glasses interaction surveys focus on wearable input and output interfaces~\cite{lee2018interaction}; and application-oriented reviews examine healthcare or industrial support~\cite{wang2025digitalhealth,danielsson2020industrialassembly}, or domain-specific acceptance~\cite{koutromanos2023acceptance}. In contrast, this survey centers on AI smart glasses as wearable-intelligence platforms for personalized assistance, emphasizing the coupling among hardware constraints, intelligence capabilities, interaction design, and real-world deployment.

\begin{tcolorbox}[agentcontrib]
This survey makes the following contributions:
\begin{itemize}
    \item \textbf{Conceptual framing:} We frame AI smart glasses as integrated wearable-intelligence platforms for personalized intelligent assistance, moving beyond views of smart glasses as sensing or display devices.

    \item \textbf{Analytical framework:} We organize AI smart glasses into four connected parts: hardware foundation, wearable intelligence, interaction design, and application scenarios, and synthesize representative work across sensing, reasoning, interaction, and deployment.

    \item \textbf{Applications and future agenda:} We connect the framework to real-world application scenarios and outline cross-cutting future directions toward personalized assistance that is adaptive, dependable, and trustworthy in wearable contexts.
\end{itemize}
\end{tcolorbox}

\clearpage
\tableofcontents
\clearpage

\section{Hardware Foundation}
\label{sec:hardware-foundation}

\begin{figure}[t]
    \centering
    \includegraphics[width=\textwidth]{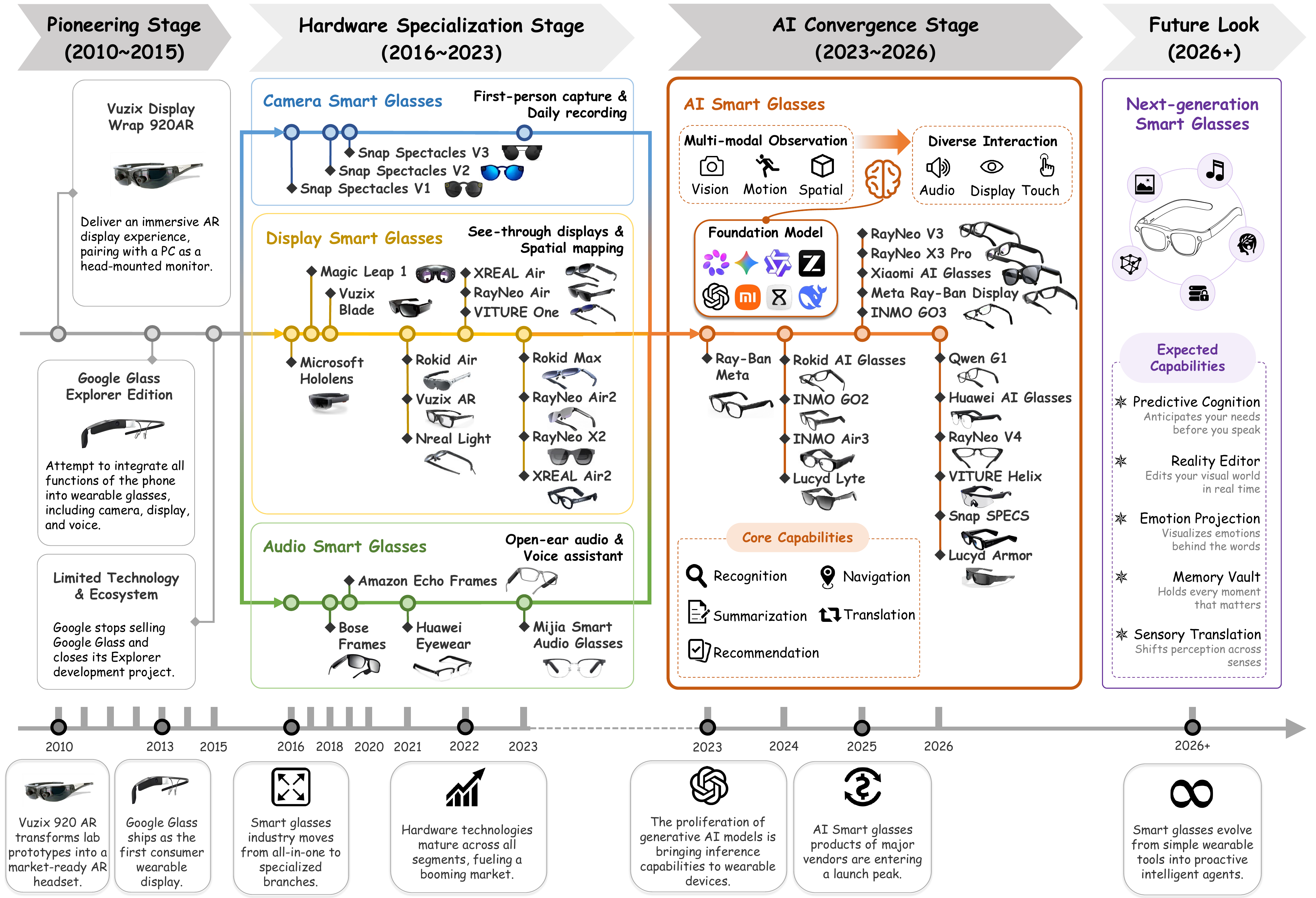}
    \caption{Evolution of smart glasses. Smart glasses have progressed from an early exploratory phase to a stage of hardware specialization and, more recently, to the convergence of wearable hardware and multimodal foundation models. Looking ahead, they are expected to evolve into proactive, embodied intelligent assistants that augment human cognition, perception, and interaction.}
    \label{fig:hardware-history}
\end{figure}

As the physical foundation of AI smart glasses, advances in wearable hardware have continuously expanded the capabilities of smart glasses over the past decade. As illustrated in Figure~\ref{fig:hardware-history}, this evolution reflects a gradual transition from simple wearable devices to AI assistants. Early attempts, represented by Vuzix and Google in the Pioneering Stage (2010-2015), pursued an all-in-one vision by integrating smartphone-like functionalities into a lightweight wearable form factor. However, limited hardware technologies and immature application ecosystems prevented this design from achieving widespread adoption. Therefore, the industry shifted toward the hardware specialization stage (2016-2023), where camera smart glasses focused on egocentric capture and daily recording, display smart glasses advanced see-through displays and spatial mapping, and audio smart glasses popularized open-ear listening and voice assistants. These specialized product lines progressively matured the underlying sensing, interaction, and display technologies that later became the hardware foundation of AI smart glasses.

The AI Convergence stage (2023-2026) has fundamentally reshaped the development trajectory, the core of which is not a simple hardware stacking, but rather a synergy between multimodal foundation models and existing hardware architecture. Unlike traditional smart glasses that could only passively record or replay information, the breakthrough of this phase lies in the closed-loop integration of multimodal perception and inference intelligence. The multimodal foundation models (ChatGPT, Gemini, Meta, Qwen, etc.) act as a central brain: they ingest multimodal observation data to perform unified semantic understanding, while orchestrating diverse interactions for comprehensive feedback. This profound synergy enables the core capabilities of AI smart glasses in recognition, navigation, summarization, translation, and recommendation. Meanwhile, these capabilities place unprecedented demands on wearable hardware, which must deliver real-time intelligence under stringent constraints on size, weight, and power consumption.

\begin{figure}[t]
    \centering
    \includegraphics[width=\textwidth]{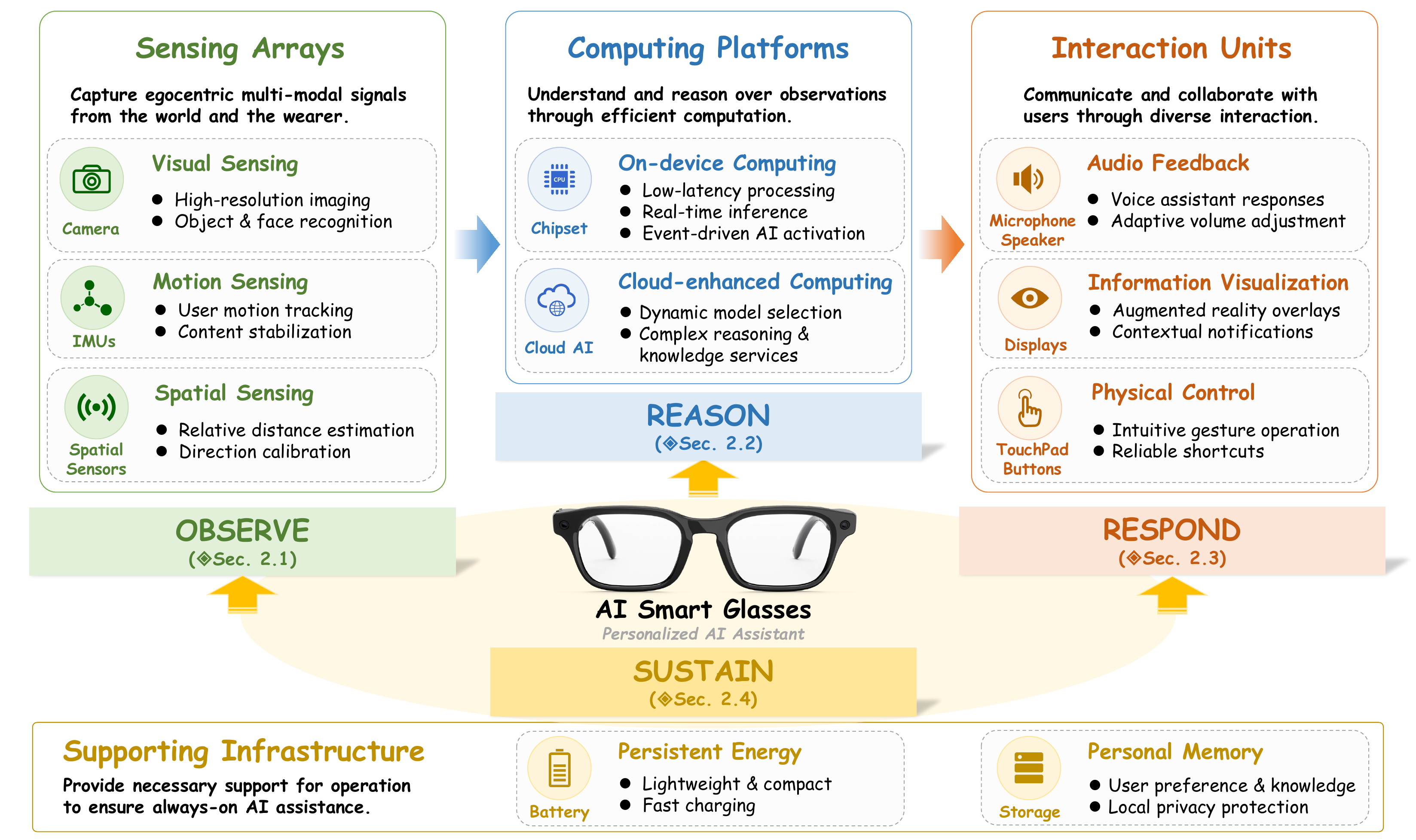}
    \caption{Overview of AI smart glasses hardware capabilities. By synthesizing the hardware components into a capability stack, AI smart glasses are able to effectively observe, reason, and respond to users in a unified workflow.}
    \label{fig:hardware-overview}
\end{figure}

From the perspective of an intelligent assistant, supporting the complete intelligence pipeline requires four fundamental hardware capabilities, as illustrated in Figure~\ref{fig:hardware-overview}. \emph{Sensing Arrays}(\S\ref{sec:sensing arrays}) continuously observe both the surrounding environment and the wearer through egocentric visual, motion, and spatial signals, providing the sensory foundation for wearable intelligence. \emph{Computing Platforms}(\S\ref{sec:computing platforms}) support contextual reasoning over heterogeneous observations by coordinating on-device processors with cloud-based AI models. \emph{Interaction Units}(\S\ref{sec:user feedback units}) establish communication channels to capture user input and present system feedback, enabling wearable assistance in mobile and attention-constrained conditions. \emph{Supporting Infrastructure}(\S\ref{sec:Supporting infrastructure}) provides reliable energy and storage, allowing smart glasses to operate reliably throughout prolonged daily use.
Rather than treating cameras, chips, controls, speakers, displays, batteries, and storage as isolated hardware components, we organize them as a progressive hardware capability stack that enables AI smart glasses to function as embodied intelligent assistants. 

Table~\ref{tab:hardware-representative} summarizes representative AI smart-glasses products and their hardware specifications along this capability stack. Most products share a common paradigm of visual sensing and multi-tiered reasoning, complemented by audio-tactile interactions, yet they differ notably in spatial awareness and display configurations. This unwavering alignment on core reasoning capabilities, however, contrasts sharply with the heterogeneous configurations in multi-sensory interaction, revealing a critical engineering trade-off between enriched user experiences and limited hardware resources. Different manufacturers balance it through distinct design philosophies and strategic prioritization, shaping the diverse product landscape.

\begin{table*}[t]
\centering
\caption{Hardware specifications of representative AI smart glasses. The dash (-) indicates that the information is not disclosed in public materials; \textit{Snapdragon AR1} and \textit{AR1} both refer to the Qualcomm Snapdragon AR1 Gen1 chipset; \textit{mic} and \textit{spk} denote microphone and speaker, respectively.}
\label{tab:hardware-representative}
\renewcommand{\arraystretch}{1.2}
\setlength{\tabcolsep}{4pt}
\resizebox{\linewidth}{!}{
\begin{tabular}{lc|ccc|cc|ccc|cc}
\toprule
\multirow{3}{*}{\textbf{Product}}
& \multirow{3}{*}{\textbf{\makecell{Release\\Time}}}
& \multicolumn{3}{c|}{\textbf{Observation}}
& \multicolumn{2}{c|}{\textbf{Reasoning}}
& \multicolumn{3}{c|}{\textbf{Responsiveness}}
& \multicolumn{2}{c}{\textbf{Sustainability}} \\

\cmidrule(lr){3-5}
\cmidrule(lr){6-7}
\cmidrule(lr){8-10}
\cmidrule(lr){11-12}

& & Camera & Motion & Spatial 
& \multirow{2}{*}{Local Inference} 
& \multirow{2}{*}{Cloud AI}
& \multirow{2}{*}{Audio}
& \multirow{2}{*}{Display}
& Touch
& \multirow{2}{*}{Battery}
& \multirow{2}{*}{Storage} \\

& & Resolution & Stabilization & Awareness 
& & & & 
& Control 
& &  \\

\midrule

Ray-Ban Meta & 2023.09
& 12 MP & $\checkmark$ & -- & Snapdragon AR1
& Meta AI
& 5-mic, 2-spk & $\times$ & $\checkmark$
& 160 mAh & 32 GB \\

\rowcolor{RowGray}
Rokid AI Glasses & 2024.11
& 12 MP & $\checkmark$ & -- & Snapdragon AR1
& Multi-LLM
& 4-mic, 2-spk & Binocular & $\checkmark$
& 210 mAh & 32 GB \\

INMO Air3 & 2024.11
& 16 MP & $\checkmark$ & $\checkmark$ & Space Computing Chip & GLM AI
& 4-mic, 2-spk & Binocular & $\checkmark$
& 660 mAh & 128 GB \\

\rowcolor{RowGray}
RayNeo X3 Pro & 2025.05
& 12 MP & $\checkmark$ & $\checkmark$ & Snapdragon AR1 & Qwen
& 3-mic, 4-spk & Binocular & $\checkmark$
& 245 mAh & 32 GB \\

Xiaomi AI Glasses & 2025.06
& 12 MP & $\checkmark$ & -- & AR1 + BES2700 & MiMo
& 5-mic, 2-spk & $\times$ & $\checkmark$
& 263 mAh & 32 GB \\

\rowcolor{RowGray}
Meta Ray-Ban Display & 2025.09
& 12 MP & $\checkmark$ & -- & Snapdragon AR1 & Meta AI
& 6-mic, 2-spk & Monocular & $\checkmark$
& 248 mAh & 32 GB \\

INMO GO3 & 2025.11
& 8 MP & $\checkmark$ & -- & Unisoc W337 & GLM AI
& 4-mic, 2-spk & Binocular & $\checkmark$
& 270 mAh & 64 GB \\

\rowcolor{RowGray}
Qwen G1 & 2026.03
& 12 MP & $\checkmark$ & -- & AR1 + BES2800 & Qwen
& 6-mic, 2-spk & $\times$ & $\checkmark$
& 272 mAh & 64 GB \\

Huawei AI Glasses & 2026.04
& 12 MP & $\checkmark$ & -- & Self-developed Chip & PanguLM
& 3-mic, 2-spk & $\times$ & $\checkmark$
& 252 mAh & 64 GB \\

\rowcolor{RowGray}
RayNeo V4 & 2026.05
& 9 MP & $\checkmark$ & -- & AR1 + BES2800BP & Qwen
& 4-mic, 2-spk & $\times$ & $\checkmark$
& 250 mAh & 64 GB \\

Snap SPECS & 2026.06
& -- & $\checkmark$ & $\checkmark$ & Dual Snapdragon Chips & -- 
& 6-mic, 2-spk & Binocular & $\checkmark$ 
& -- & -- \\

\rowcolor{RowGray}
VITURE Helix & 2026.06 
& 12 MP & -- & -- & -- & NVIDIA XR AI 
& 4-mic, 2-spk & $\times$ & $\checkmark$ 
& -- & -- \\

\bottomrule
\end{tabular}
}
\end{table*}

\subsection{Sensing Arrays}
\label{sec:sensing arrays}
Sensing arrays serve as a gateway through which AI smart glasses perceive the physical world. Unlike traditional wearable devices that collect isolated physiological or motion signals, AI smart glasses continuously observe both the environment and the wearer from an egocentric perspective to support downstream reasoning and response. By capturing visual scenes, user motions, and surrounding spatial structures, sensing arrays establish the hardware basis for transforming raw sensory inputs into wearable intelligence.

Visual sensing provides the primary means of acquiring environmental information for smart glasses. Most commercial devices are equipped with ultra-wide cameras ranging from 12 to 16 megapixels, providing sufficient visual fidelity for object recognition, scene observation, and activity recording~\cite{meta2025}. Continuous egocentric imaging enables AI assistants to observe what users see and establish a shared visual context for subsequent reasoning. Recent products further improve imaging quality through High Dynamic Range (HDR) imaging and low-light enhancement, preserving image fidelity under challenging illumination conditions~\cite {huawei2026}. Beyond environmental observation, some visual sensors support face recognition and virtual contact card generation, enabling social interaction assistance~\cite{inmo2025}.

Complementing visual perception, Inertial Measurement Units (IMUs) enable smart glasses to detect the wearer's movements. This dynamic information allows the system to stabilize captured imagery and virtual content while providing important cues about user behaviors and attention~\cite{xreal2024}. Motion sensing also supports low-latency interaction by detecting head gestures and body movements, thereby complementing interaction modalities~\cite{rokid2023}. As AI assistants become increasingly proactive, continuous motion awareness provides essential behavioral context for understanding user intentions and activities throughout daily life.

Spatial sensing further extends environment observation to a 3D scene. By integrating depth sensors, distance sensors, and magnetometers, modern smart-glasses estimate distances and calibrate object orientations within a scene, thereby constructing rich spatial signals~\cite{rayneo2025}. The fusion of diverse signals enables AI assistants to understand the wearer's surroundings, providing the foundational infrastructure for real-time spatial localization and navigation assistance.

\subsection{Computing Platforms}
\label{sec:computing platforms}
Building on the multimodal observations captured by sensing arrays, computing platforms provide the processing foundation that enables AI smart glasses to interpret their surroundings and support user assistance. 
In practice, this processing foundation can be implemented through different computing modes, ranging from on-device computing for low-latency, privacy-sensitive inference to cloud-enhanced computing for scalable model capacity and service integration.

Modern smart glasses employ highly integrated on-device computing to enable real-time intelligence inference under stringent resource constraints. 
The System-on-Chip (SoC) integrates Central Processing Units (CPUs), Graphics Processing Units (GPUs), Neural Processing Units (NPUs), and Image Signal Processors (ISPs) into a unified platform, enabling efficient execution of sensor processing and neural inference workloads.
Representative platforms such as the Qualcomm Snapdragon AR1 Gen 1 further optimize this integration for visual understanding, speech recognition, and multimodal feature extraction on edge devices~\cite{snapdragon2023}. The tightly coupled design reduces inference latency and improves energy efficiency, making it the dominant computing solution in recent commercial smart glasses.

To further trade off computational capability with energy efficiency, some systems employ hierarchical on-device computing architectures. Instead of executing all workloads on a single processor, these architectures separate always-on sensing from computationally intensive reasoning. An ultra-low-power microcontroller unit (MCU) continuously monitors wake words, inertial measurements, and other lightweight sensing signals, while the primary processor remains dormant until user intent or environmental events are detected. This event-driven scheduling reduces unnecessary computation and extends operating time without sacrificing responsiveness. Commercial products such as Xiaomi AI Glasses~\cite{xiaomi2025} and RayNeo V4~\cite{rayneo2025} have adopted dual-chip architectures that combine low-power control processors with high-performance AI chips, demonstrating that hierarchical computing resource management has become an increasingly practical strategy for continuous wearable intelligence.

Despite the development of on-device chips, the computational demands of multimodal foundation models still exceed the capabilities of current wearable on-device processors. Consequently, cloud-enhanced computing has become an indispensable component for smart glasses. AI assistants typically perform lightweight inference locally, while computationally intensive reasoning is delegated to cloud-hosted foundation models. For instance, Rokid AI Glasses support dynamic switching among multiple AI models, including DeepSeek, Qwen, and GLM AI, allowing users to select specialized reasoning capabilities for different tasks~\cite{rokid2024}. This collaborative architecture not only overcomes hardware computational limitations but also enables smart glasses to leverage rapidly evolving large multimodal models without frequent hardware upgrades.

\subsection{Interaction Units}
\label{sec:user feedback units}

Interaction units provide the physical input and output channels through which AI smart glasses connect computational results with the wearer. While sensing arrays observe the physical world and computing platforms convert observations into task-relevant outputs, interaction units determine how user input can be captured and how system feedback can be presented under mobile and attention-constrained conditions. By combining acoustic, visual, and physical channels, they allow AI assistance to remain continuously accessible while reducing the cognitive and operational burden of device access.

Audio interaction has become the primary communication channel for AI smart glasses. Microphone arrays capture user speech for voice commands and conversation content, while open-ear speakers deliver AI-generated responses and multimedia notifications. Recent audio devices improve communication quality through beamforming, adaptive volume adjustment, far-field speech enhancement, and acoustic leakage suppression, allowing AI assistants to support natural conversations across diverse acoustic environments~\cite{rayneo2026}. 
As large multimodal models reshape human-AI interaction, speech is increasingly evolving into the most intuitive interface between humans and wearable AI.

Visual feedback complements audio interaction by presenting information-rich content directly on the lenses. Rather than requiring users to shift their attention toward handheld devices, smart glasses overlay digital information onto the visual field, enabling glanceable access to navigation cues, translation subtitles, and contextual notifications. Commercial systems such as Meta Ray-Ban Display and INMO GO series employ Micro-LED displays and diffractive waveguide optics to achieve lightweight optical designs while maintaining sufficient brightness and readability for outdoor use~\cite{inmo2025, inmo2024}. Compared with transient voice responses, visual feedback allows AI assistants to communicate complex information continuously without interrupting ongoing activities, making it particularly valuable for navigation and multilingual communication assistance.

Physical interaction provides reliable and low-latency control when voice communication is unavailable or undesirable. A capacitive touchpad integrated into the temples supports intuitive gestures, such as tapping and swiping, for assistant activation and media control. Meanwhile, mechanical buttons provide reliable shortcuts for photography, recording, translation, and other latency-sensitive functions. These lightweight input mechanisms further complement audio interaction by enabling users to communicate explicit intentions in noisy environments or scenarios requiring rapid operation.

\subsection{Supporting Infrastructure}
\label{sec:Supporting infrastructure}
Supporting infrastructure forms the foundational layer that enables AI smart glasses to function as persistent intelligent assistants in real-world environments. Without a reliable energy supply and data storage, the capabilities of observation, reasoning, and response cannot be sustained. 
Consequently, supporting infrastructure determines both the operating duration of AI services and the reliable storage of personal and contextual data under the stringent size, weight, and power constraints of wearable devices.

Among these supporting components, energy supply is the most immediate constraint. Despite integrating cameras, microphones, processors, and displays, most commercial smart glasses are powered by batteries with capacities of only 200-300mAh, representing less than one twentieth of the battery capacity of a typical smartphone. The limited energy budgets require every stage of sensing, computation, and communication to be carefully optimized for power efficiency. Rather than simply increasing battery capacity, recent devices improve service continuity through system-level energy management, including low-power computing architectures ~\cite{xiaomi2025}, event-driven task scheduling ~\cite{rayneo2025}, and portable charging cases that enable additional fast-charging cycles ~\cite{rokid2024}. They enable users to replenish energy opportunistically during daily routines, extending the effective operating duration of AI assistance from several hours to an entire day.

Personal data storage also plays a crucial role in enabling personalized intelligence throughout long-term human-AI interactions. Unlike conventional storage systems that primarily preserve application resources and multimedia files, the storage of AI smart glasses is a repository for user-specific knowledge, including local personalized AI models, personal preferences, interaction histories, and contextual records accumulated during daily use. By retaining and updating these personalized memories, AI assistants can build a deeper understanding of users and deliver context-aware, personalized support over time. Meanwhile, personal memory also strengthens privacy preservation. Sensitive information, such as facial features, voiceprints, and frequently accessed personal data, can be processed and retained locally without transmission to cloud servers, reducing privacy risks while maintaining continuous personalized services. 
As foundation models become increasingly capable of long-term memory and personalized reasoning, future smart glasses are expected to rely more heavily on on-device storage to preserve persistent user profiles and contextual knowledge across extended interactions. Consequently, storage systems are evolving from passive data repositories into hardware substrates that enable continuous personalization and long-term intelligent assistance.

\textbf{Outlook.}
Looking beyond current AI smart glasses, next-generation systems are expected to support more persistent, contextual, and proactive forms of wearable intelligence. The capabilities summarized in Figure~\ref{fig:hardware-history}, including predictive cognition, real-time perceptual augmentation, emotion-aware communication, persistent personal memory, and cross-sensory translation, illustrate how future assistance may extend from explicit responses to continuous situated support. From a hardware perspective, these capabilities intensify requirements across the entire capability stack: multimodal sensing must capture richer egocentric, behavioral, spatial, and social signals; heterogeneous computing must coordinate local inference and cloud-enhanced reasoning under latency and energy constraints; interaction units must deliver timely and controllable feedback without overwhelming attention; and supporting infrastructure must sustain long-term operation and personal data management. Therefore, the hardware foundation of AI smart glasses should be understood not only as a collection of components, but as the enabling substrate that bounds how wearable intelligence can be realized in daily environments.

\section{Wearable Intelligence}
\label{sec:wearable-intelligence}
\tikzstyle{my-box}=[
rectangle,
draw=DarkBlue,
rounded corners,
text opacity=1,
minimum height=1.5em,
minimum width=5em,
inner sep=2pt,
align=center,
fill opacity=.5,
]
\tikzstyle{leaf}=[
my-box,
fill=yellow!32,
text=black,
align=center,
font=\small,
inner xsep=5pt,
inner ysep=4pt,
text width=38em,
]
\tikzstyle{leaf2}=[
leaf,
fill=purple!20,
]
\tikzstyle{leaf3}=[
leaf,
fill=LightBlue,
]
\newcommand{\taxnode}[1]{\makebox[9em][c]{\shortstack[c]{#1}}}
\newcommand{\taxleaf}[1]{\begin{minipage}{38em}\raggedright #1\end{minipage}}

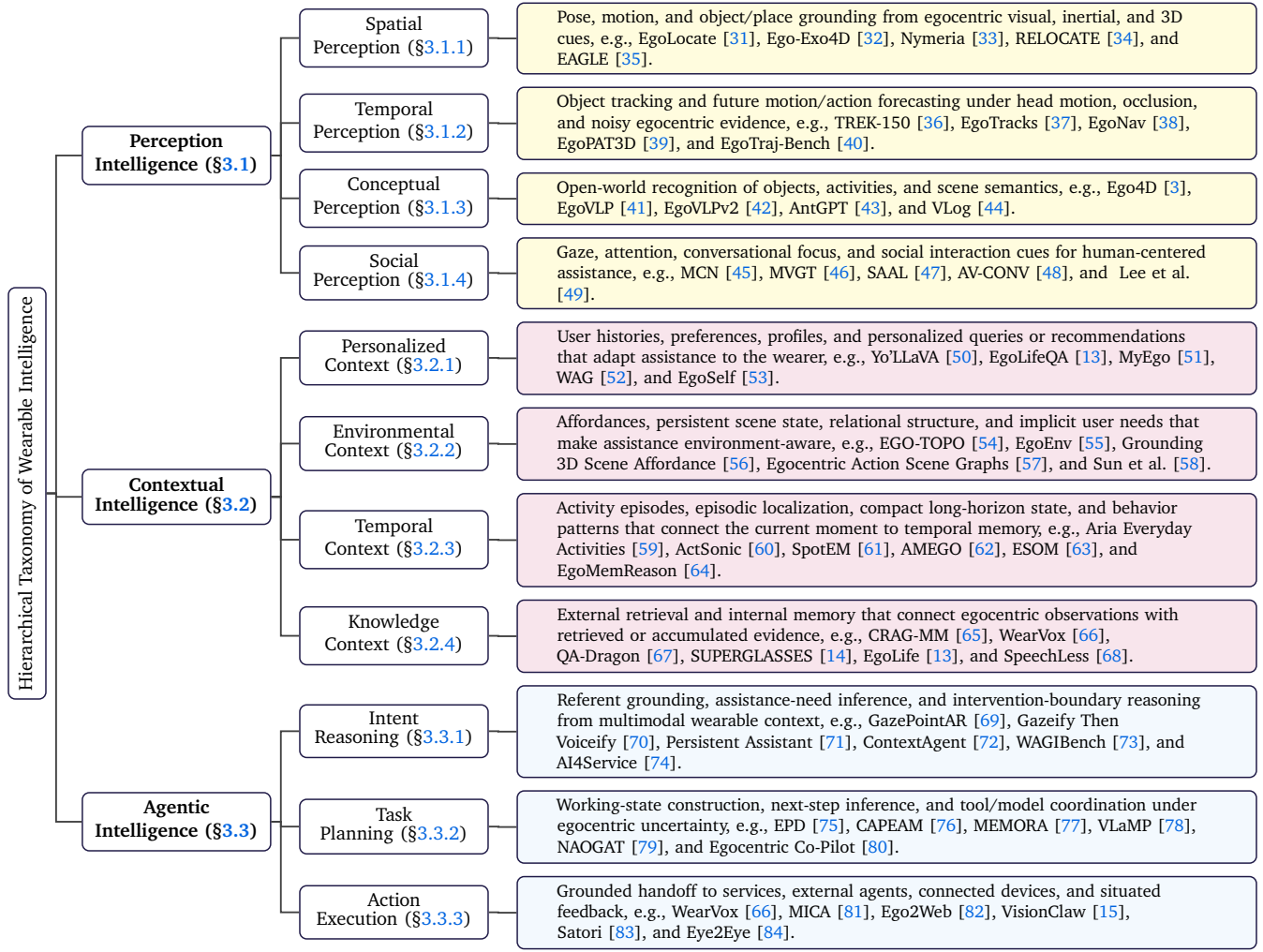
\begin{figure*}[t]
\vspace{-2mm}
\centering
\resizebox{\textwidth}{!}{
    \begin{forest}
        forked edges,
        for tree={
            grow=east,
            reversed=true,
            anchor=base west,
            parent anchor=east,
            child anchor=west,
            base=left,
            font=\normalsize,
            rectangle,
            draw=DarkBlue,
            rounded corners,
            align=center,
            minimum width=4em,
            edge+={darkgray, line width=1pt},
            s sep=7pt,
            l sep=16pt,
            inner xsep=5pt,
            inner ysep=4pt,
            line width=.8pt,
            ver/.style={rotate=90, child anchor=north, parent anchor=south, anchor=center},
        },
        where level=1{text width=9em,font=\normalsize\bfseries,align=center}{},
        where level=2{text width=9em,font=\normalsize,align=center}{},
        where level=3{text width=38em,font=\small,align=left}{},
[Hierarchical Taxonomy of Wearable Intelligence, ver
    [\taxnode{Perception\\Intelligence (\S\ref{sec:perception-intelligence})}
        [\taxnode{Spatial\\Perception (\S\ref{sec:spatial-perception})}
            [{\taxleaf{Pose, motion, and object/place grounding from egocentric visual, inertial, and 3D cues, e.g., EgoLocate~\cite{yi2023egolocate}, Ego-Exo4D~\cite{grauman2024ego}, Nymeria~\cite{ma2024nymeria}, RELOCATE~\cite{khosla2025relocate}, and EAGLE~\cite{cao2026eagle}.}}, leaf]
        ]
        [\taxnode{Temporal\\Perception (\S\ref{sec:temporal-perception})}
            [{\taxleaf{Object tracking and future motion/action forecasting under head motion, occlusion, and noisy egocentric evidence, e.g., TREK-150~\cite{dunnhofer2023visual}, EgoTracks~\cite{tang2023egotracks}, EgoNav~\cite{wang2024egonav}, EgoPAT3D~\cite{li2022egocentric}, and EgoTraj-Bench~\cite{liu2025egotraj}.}}, leaf]
        ]
        [\taxnode{Conceptual\\Perception (\S\ref{sec:conceptual-perception})}
            [{\taxleaf{Open-world recognition of objects, activities, and scene semantics, e.g., Ego4D~\cite{grauman2022ego4d}, EgoVLP~\cite{lin2022egocentric}, EgoVLPv2~\cite{pramanick2023egovlpv2}, AntGPT~\cite{zhao2024antgpt}, and VLog~\cite{lin2025vlog}.}}, leaf]
        ]
        [\taxnode{Social\\Perception (\S\ref{sec:social-perception})}
            [{\taxleaf{Gaze, attention, conversational focus, and social interaction cues for human-centered assistance, e.g., MCN~\cite{huang2020mutual}, MVGT~\cite{miao2025multi}, SAAL~\cite{ryan2023egocentric}, AV-CONV~\cite{jia2024audio}, and ~\citet{lee2024modeling}.}}, leaf]
        ]
    ]
    [\taxnode{Contextual\\Intelligence (\S\ref{sec:contextual-intelligence})}
        [\taxnode{Personalized\\Context (\S\ref{sec:personal-context})}
            [{\taxleaf{User histories, preferences, profiles, and personalized queries or recommendations that adapt assistance to the wearer, e.g., Yo'LLaVA~\cite{nguyen2024yo}, EgoLifeQA~\cite{yang2025egolife}, MyEgo~\cite{xiao2026ego}, WAG~\cite{lu2026query}, and EgoSelf~\cite{wang2026egoself}.}}, leaf2]
        ]
        [\taxnode{Environmental\\Context (\S\ref{sec:environmental-context})}
            [{\taxleaf{Affordances, persistent scene state, relational structure, and implicit user needs that make assistance environment-aware, e.g., EGO-TOPO~\cite{nagarajan2020egotopo}, EgoEnv~\cite{nagarajan2023egoenv}, Grounding 3D Scene Affordance~\cite{liu2024grounding}, Egocentric Action Scene Graphs~\cite{rodin2024action}, and \citet{sun2025visual}.}}, leaf2]
        ]
        [\taxnode{Temporal\\Context (\S\ref{sec:temporal-context})}
            [{\taxleaf{Activity episodes, episodic localization, compact long-horizon state, and behavior patterns that connect the current moment to temporal memory, e.g., Aria Everyday Activities~\cite{lv2024aria}, ActSonic~\cite{mahmud2024actsonic}, SpotEM~\cite{ramakrishnan2023spotem}, AMEGO~\cite{goletto2024amego}, ESOM~\cite{manigrasso2026online}, and EgoMemReason~\cite{wang2026egomemreason}.}}, leaf2]
        ]
        [\taxnode{Knowledge\\Context (\S\ref{sec:knowledge-context})}
            [{\taxleaf{External retrieval and internal memory that connect egocentric observations with retrieved or accumulated evidence, e.g., CRAG-MM~\cite{wang2025cragmm}, WearVox~\cite{lin2025wearvox}, QA-Dragon~\cite{jiang2025qa}, SUPERGLASSES~\cite{jiang2026superglasses}, EgoLife~\cite{yang2025egolife}, and SpeechLess~\cite{kim2026speechless}.}}, leaf2]
        ]
    ]
    [\taxnode{Agentic\\Intelligence (\S\ref{sec:agentic-intelligence})}
        [\taxnode{Intent\\Reasoning (\S\ref{sec:intent-reasoning})}
            [{\taxleaf{Referent grounding, assistance-need inference, and intervention-boundary reasoning from multimodal wearable context, e.g., GazePointAR~\cite{lee2024gazepointar}, Gazeify Then Voiceify~\cite{zhang2026gazeify}, Persistent Assistant~\cite{cho2025persistent}, ContextAgent~\cite{yang2025contextagent}, WAGIBench~\cite{veerabadran2025benchmarking}, and AI4Service~\cite{wen2025service}.}}, leaf3]
        ]
        [\taxnode{Task\\Planning (\S\ref{sec:task-planning})}
            [{\taxleaf{Working-state construction, next-step inference, and tool/model coordination under egocentric uncertainty, e.g., EPD~\cite{shi2024epd}, CAPEAM~\cite{kim2023context}, MEMORA~\cite{yu2026memora}, VLaMP~\cite{patel2023pretrained}, NAOGAT~\cite{thakur2024leveraging}, and Egocentric Co-Pilot~\cite{yang2026egocentric}.}}, leaf3]
        ]
        [\taxnode{Action\\Execution (\S\ref{sec:action-execution})}
            [{\taxleaf{Grounded handoff to services, external agents, connected devices, and situated feedback, e.g., WearVox~\cite{lin2025wearvox}, MICA~\cite{sarch2025grounding}, Ego2Web~\cite{yu2026ego2web}, VisionClaw~\cite{liu2026visionclaw}, Satori~\cite{li2025satori}, and Eye2Eye~\cite{teng2026seeing}.}}, leaf3]
        ]
    ]
]
    \end{forest}
}
\caption{Taxonomy of wearable intelligence for AI smart glasses, consisting of perception, contextual, and agentic intelligence.}
\label{fig:taxonomy}
\end{figure*}

Unlike traditional AI systems that primarily perceive the world from third-person observations, smart glasses require a continuous egocentric perspective to provide actionable assistance within the constraints of wearable devices. We refer to this capability as \emph{Wearable Intelligence}: the progressive conversion of egocentric observations into intelligent support.
In this survey, we organize wearable intelligence into three hierarchical levels, as illustrated in Figure~\ref{fig:taxonomy}. \emph{Perception Intelligence}(\S\ref{sec:perception-intelligence}) extracts directly observable information from egocentric multimodal streams, including spatial structures, temporal dynamics, semantic concepts, and social interactions.  \emph{Contextual Intelligence}(\S\ref{sec:contextual-intelligence}) enriches these cues with personal history, environmental context, temporal continuity, and external knowledge, enabling context-aware understanding. Finally, \emph{Agentic Intelligence}(\S\ref{sec:agentic-intelligence}) further turns contextual understanding into goal-directed assistance through intent reasoning, task planning, and action execution. The three levels form a progressive intelligence hierarchy that moves from observation to understanding and ultimately to assistance, illustrating how AI smart glasses evolve from passive sensing devices into proactive wearable assistants.

\subsection{Perception Intelligence}
\label{sec:perception-intelligence}

\begin{figure}[t]
    \centering
    \includegraphics[width=\textwidth]{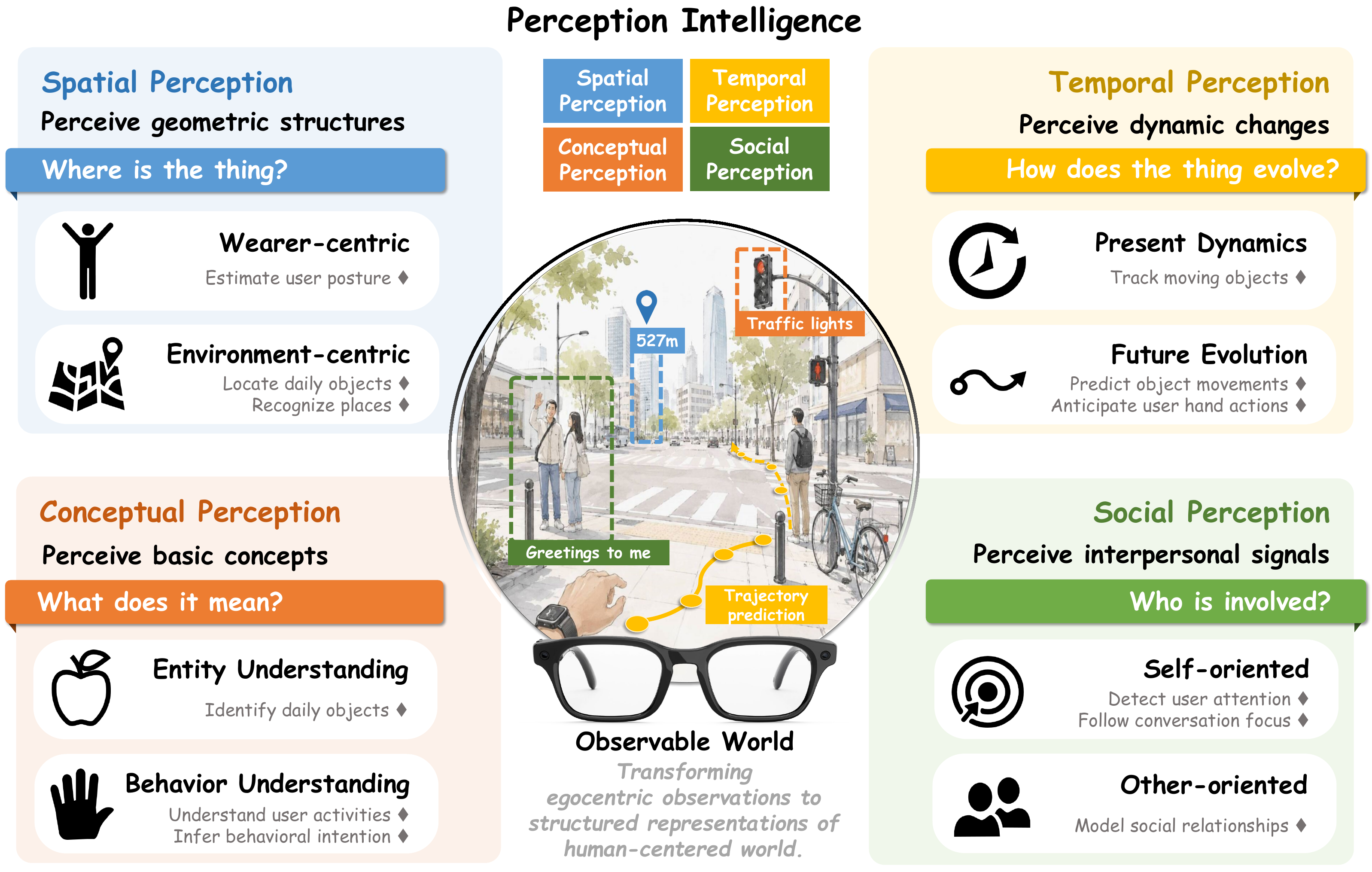}
    \caption{Overview of perception intelligence in AI smart glasses. Continuous egocentric sensing is organized into spatial, temporal, conceptual, and social perception, which respectively ground geometric relationships, dynamic evolution, semantic entities and behaviors, and interpersonal signals for downstream contextual and agentic intelligence.}
    \label{fig:perception-intelligence}
\end{figure}

Perception intelligence enables smart glasses to extract information directly from egocentric observations. As the entry point to the intelligence hierarchy, perception intelligence transforms raw sensory signals into structured representations that support subsequent contextual understanding and agentic decision-making.
In this survey, we organize perception intelligence into four interconnected capabilities, as illustrated in Figure~\ref{fig:perception-intelligence}. \emph{Spatial perception} (\S\ref{sec:spatial-perception}) captures geometric structures and spatial relationships within the physical world. \emph{Temporal perception} (\S\ref{sec:temporal-perception}) characterizes dynamic changes and future evolution from continuous egocentric observations. \emph{Conceptual perception} (\S\ref{sec:conceptual-perception}) recognizes meaningful entities and behaviors from egocentric experiences. Finally, \emph{social perception} (\S\ref{sec:social-perception}) understands interpersonal attention and social interactions in human-centered environments. Together, these capabilities enable smart glasses to answer where things are, how the world evolves, what things mean, and who people interact with, providing complementary perspectives for AI smart glasses.

\subsubsection{Spatial Perception}
\label{sec:spatial-perception}
Spatial perception in smart glasses refers to the ability to understand geometric structures and spatial relationships from an egocentric perspective. It can be categorized into two complementary forms: wearer-centric spatial perception estimates the user's body states and poses, while environment-centric spatial perception captures the layout of nearby objects and the user's position in the environment.

Wearer-centric spatial perception provides the geometric foundation for understanding user behaviors. Early methods~\cite{jiang2021egocentric} relied primarily on single-modality visual inputs for pose estimation; EgoEgo~\cite{li2023ego} recovered full-body motion from head movement, demonstrating the feasibility of egocentric pose estimation. However, image-based approaches suffer from self-occlusion, rapid ego-motion, and viewpoint transitions. Recent work~\cite{yi2023egolocate, guzov2025hmd} increasingly fuses visual observations with body-worn IMU signals to improve robustness. This shift has been further supported by large-scale multimodal egocentric datasets, including Ego-Exo4D~\cite{grauman2024ego}, Nymeria~\cite{ma2024nymeria}, and SimXR~\cite{luo2024real}, which provide synchronized egocentric video, IMU signals, and motion annotations with Project Aria glasses. EMHI~\cite{fan2025emhi} demonstrates that fusing stereo vision and IMU signals mitigates self-occlusion and drift, while Ego4o~\cite{wang2025ego4o} further supports flexible combinations of sparse IMU signals, visual streams, and motion descriptions, enabling mutual enhancement between spatial perception and embodied understanding.

Environment-centric spatial perception encompasses two main directions: object localization and place localization. Object localization identifies and localizes target objects in the current scene, which is particularly useful for wearable assistance, such as locating misplaced items. Early methods based on feature matching and region proposals struggled under severe viewpoint variations and cluttered environments of smart glasses. To address these limitations, CoCoFormer~\cite{xu2023my} improves accuracy by constructing viewpoint transformations in contrastive learning, and PRVQL~\cite{fan2025prvql} introduces a feature knowledge base to reduce background interference. More recently, RELOCATE~\cite{khosla2025relocate} leverages pretrained representations for robust spatial reasoning, while EAGLE~\cite{cao2026eagle} unifies 2D and 3D localization for consistent spatial grounding. Place localization estimates the user's location by matching observations against reference images, which is crucial for continuous navigation assistance and environment-aware services across indoor and outdoor scenarios. R$^2$Former~\cite{zhu2023r2former} combines global scene representations with relational reasoning for stable environmental awareness, and SemVPR~\cite{zhang2025efficient} further improves efficiency for real-time navigation deployment.

\subsubsection{Temporal Perception}
\label{sec:temporal-perception}
Beyond 3D spatial structure, the real world also evolves along the temporal dimension, which encodes information about the past, present, and future states. For smart glasses, temporal perception refers to the capability of continuously capturing dynamic changes and predictive reasoning from continuous egocentric observations. It enables real-time assistance and proactive warning during daily activities. We organize recent research into two directions: present object tracking and future trajectory prediction in egocentric vision.

Egocentric object tracking is the foundation of understanding dynamic changes by maintaining the temporal consistency of user-scene interactions. Compared with third-person tracking, tracking in smart glasses is substantially more challenging due to head motion, rapid locomotion, and hand occlusions. To systematically evaluate these challenges, TREK-150~\cite{dunnhofer2023visual} establishes a benchmark for temporal tracking in egocentric settings, while EgoTracks~\cite{tang2023egotracks} extends evaluation to long-duration videos and reveals difficulties unique to long-term tracking. Beyond 2D tracking, 3D object tracking provides spatial grounding for path planning and navigation. IT3DEgo~\cite{zhao2024instance} leverages camera poses and depth information to project 2D instances into 3D space, enabling persistent tracking for smart glasses in real-world scenarios.

Future trajectory prediction requires smart glasses to reason about future motion evolution from historical observations, enabling proactive assistance and hazard warnings. EgoNav~\cite{wang2024egonav} generates multiple plausible future paths to address uncertainty in navigation, while HOIMotion~\cite{hu2024hoimotion} extends prediction to body parts interacting with manipulated objects by leveraging body poses and object states. However, trajectory prediction in smart glasses faces noise problems, as historical observations are not entirely reliable. EgoTraj-Bench~\cite{liu2025egotraj} addresses this issue by constructing a real-world benchmark that reveals a substantial gap between offline evaluation and real-world deployment. Forecasting hand trajectories is particularly important for proactive interaction and intent understanding. Ego4D~\cite{grauman2022ego4d} introduces an egocentric benchmark for continuous 2D hand coordinate prediction, and EgoPAT3D~\cite{li2022egocentric} extends it to 3D by forecasting hand interaction targets. While trajectories may indicate targets, uncertain motions may also indicate hesitation or decision-making, making hand trajectory prediction not merely coordinate regression but a window into future user intentions. USST~\cite{bao2023uncertainty} infers user planning behaviors from 3D hand trajectories, while MADiff~\cite{ma2025madiff} incorporates hand-scene interactions with motion signals of smart glasses to guide future action reasoning.

\subsubsection{Conceptual Perception}
\label{sec:conceptual-perception}
Conceptual perception in smart glasses refers to the capability of interpreting the meaning of entities and behaviors from egocentric observations. It enables the system to recognize meaningful objects, interpret user behaviors, and anticipate activities, providing the semantic foundation for higher-level contextual intelligence. We organize this perception into two directions: daily entity understanding and human behavior understanding.

Daily entity understanding enables smart glasses to associate visual objects with concepts and determine their categories, supporting object search, environment-aware interaction, and daily activities. Large-scale egocentric datasets such as Ego4D~\cite{grauman2022ego4d} provide benchmarks for object understanding under real-world egocentric settings. Early methods relied on closed-set learning with predefined categories~\cite{krizhevsky2017imagenet, liu2023efficientvit}, which generalize poorly to open-world environments with unseen objects. Recent vision-language pretraining has gained significant attention for its ability to leverage large-scale internet data~\cite{lin2024vila, wang2026time, liu2026kangaroo, xu2026egodtm}. EgoVLP~\cite{lin2022egocentric} and EgoVLPv2~\cite{pramanick2023egovlpv2} explore this paradigm in egocentric video understanding, aligning observations with flexible descriptions rather than fixed categories and improving the open-world object recognition capability of smart glasses.

Understanding human behavior marks the beginning of user activity perception in smart glasses. Compared with object recognition, behavior understanding introduces stronger temporal and interaction dynamics into conceptual perception and directly supports context-aware assistance. Early methods adapted general action recognition models~\cite{feichtenhofer2019slowfast, wu2022memvit, patrick2021keeping} to egocentric datasets such as Ego4D~\cite{grauman2022ego4d}, EPIC-Kitchens-100~\cite{damen2022rescaling}, and Assembly101~\cite{sener2022assembly101}, enabling smart glasses systems to recognize user activities. However, these approaches typically rely on predefined action vocabularies, limiting their generalization to diverse real-world behaviors and interaction patterns. Smart glasses instead require open-world understanding of behavior and state, as human activities and affective cues vary across environments, tasks, and sensing modalities~\cite{wang2024multimodal, quan2025semantic, he2026cross, huang2026grading}. To address this limitation, several studies have adopted multimodal models with zero-shot generalization capabilities~\cite{kim2025openvla, qu2024llms}. ActionCLIP~\cite{wang2023actionclip} transfers semantic knowledge from multimodal models to egocentric action recognition, enabling the identification of unseen actions without task-specific annotations. OAP~\cite{chatterjee2023opening} further models the relationship between user actions and manipulated objects, improving the semantic generalization of wearable systems across interaction scenarios. 

Recent work extends behavior understanding from recognizing current actions to anticipating future activities and interaction intentions~\cite{girase2023latency, mittal2024can, guo2024uncertainty}. AntGPT~\cite{zhao2024antgpt} leverages LLMs for goal-oriented reasoning over long-term action representations to predict future behaviors from egocentric activity streams, while VLog~\cite{lin2025vlog} expands the semantic vocabulary of novel events for open-world anticipation in smart glasses systems. By semantically predicting what users are likely to do next, smart glasses can establish the foundation for task planning and anticipatory interaction.

\subsubsection{Social Perception} 
\label{sec:social-perception}
Social perception in smart glasses refers to the ability to understand interpersonal signals from egocentric observations. This capability is particularly important for smart glasses to provide effective social assistance and communication support in real-world settings. According to the social dimension, we organize it into two directions: self-oriented social perception and other-oriented social perception.

Estimating the wearer's attention is foundational for self-oriented social perception. By inferring gaze direction, smart glasses identify which objects or individuals currently attract attention, providing clues about user intentions and social engagement. MCN~\cite{huang2020mutual} demonstrates that modeling action distributions significantly improves attention prediction performance, because hand actions and gaze behaviors are strongly correlated. MVGT~\cite{miao2025multi} further leverages multi-view observations~\cite{he2025dual,he2026prototype} to capture eye movements and attended targets, improving gaze estimation under dynamic interactions. More recently, smart glasses with integrated eye-tracking sensors enable direct estimation of gaze directions and attention distributions, providing reliable measurements in natural environments~\cite{meyer2025ambient, ma2025mmet}.

Social attention estimation is critical for conversational assistance, where smart glasses determine the intended speaker among multiple participants. Grounding user attention to specific partners enables selective speech amplification and noise suppression, improving communication quality in crowded environments~\cite{jiang2022egocentric, xue2023egocentric}. SAAL~\cite{ryan2023egocentric} pioneers this direction by jointly leveraging egocentric video and multichannel audio signals to infer attended speakers in multi-party conversations. Recent work AV-CONV~\cite{jia2024audio} extends attention estimation toward conversational focus understanding, enabling continuous tracking of shifts in conversational attention.

Other-oriented social perception treats individuals as social agents and models their behaviors and conversations, enabling smart glasses to perceive social structures and maintain awareness of multiple participants. To facilitate social modeling, recent benchmarks span dialogue understanding~\cite{chang2026multimodal, lee2024modeling}, social video question answering~\cite{lei2018tvqa, mathur2025social,hyun2024smile}, and social behavior recognition~\cite{cao2025socialgesture}, evaluating whether systems can infer latent social relationships and interaction structures from multimodal observations. Recent multimodal large language models further improve the integration of visual and auditory social cues for smart glasses, enabling more reliable identification of conversational turns and interaction targets~\cite{zhang2026can}. Aligning attention across multiple participants further improves robustness of social perception, enabling better understanding of social relationships in multi-person environments~\cite{ouyang2026multi}.

\textbf{Summary.}
These four forms of perception enable smart glasses to transform raw egocentric observations into structured representations of the human-centered world, thereby answering what is directly observable. The central challenge is unified multimodal continuity: maintaining coherent representations under wearable constraints while balancing accuracy and robustness. Evaluation should therefore move beyond task-specific accuracy to consider perceptual continuity, multimodal consistency, computational efficiency, and robustness in real-world deployments.

\subsection{Contextual Intelligence}
\label{sec:contextual-intelligence}

\begin{figure}
    \centering
    \includegraphics[width=\textwidth]{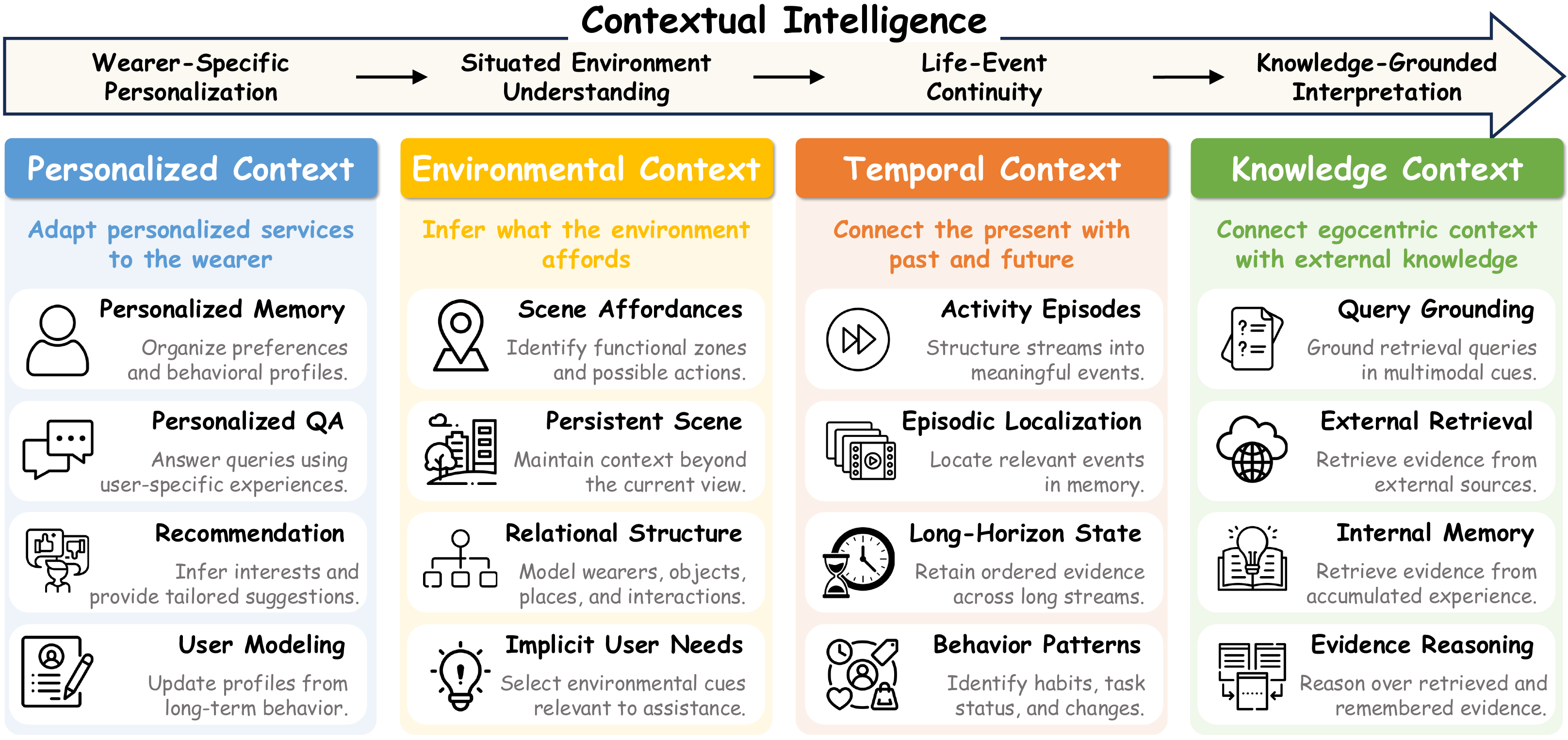}
    \caption{Overview of contextual intelligence in AI smart glasses. Egocentric observations are interpreted through personalized, environmental, temporal, and knowledge context, allowing the system to relate perceived cues to the wearer's profile, surrounding situation, life-event continuity, external and internal knowledge.}
    \label{fig:contextual-intelligence}
\end{figure}

Contextual intelligence extends perception intelligence by interpreting egocentric observations in relation to the wearer, the surrounding situation, temporal continuity, and relevant knowledge. As illustrated in Figure~\ref{fig:contextual-intelligence}, we organize this capability into four complementary forms: \emph{personalized context} (\S\ref{sec:personal-context}), which adapts interpretation to user histories and preferences; \emph{environmental context} (\S\ref{sec:environmental-context}), which abstracts the surrounding scene into affordances and situational meanings; \emph{temporal context} (\S\ref{sec:temporal-context}), which links the current moment to prior episodes, routines, and future needs; and \emph{knowledge context} (\S\ref{sec:knowledge-context}), which connects egocentric observations to external retrieval and personal memory.

\subsubsection{Personalized Context}
\label{sec:personal-context}
Personalized context incorporates user histories, preferences, routines, and profiles into contextual reasoning, enabling the same observation to be interpreted differently across wearers. Such capability is essential for AI smart glasses to evolve from generic perception systems into personalized assistants that can provide user-specific support in daily life. Existing research mainly explores three directions: personalized memory, personalized question answering, and personalized recommendation.

Personalized memory organizes user-specific preferences, habits, and behavioral profiles into structured personal knowledge. Yo'LLaVA~\cite {nguyen2024yo} directly injects personal visual information as contextual signals for user-specific concept understanding and visual reasoning. MemoryBank~\cite{zhong2024memorybank} introduces storage and retrieval modules for long-term personality modeling, while Generative Agents~\cite{park2023generative} represent personal experiences in natural language and retrieve relevant memories for behavioral prediction and planning. Because accumulated user histories can become redundant and inefficient to retrieve, recent work has shifted from memory storage to memory organization and abstraction. AMEGO~\cite{goletto2024amego} organizes user interactions captured from egocentric videos into active memory representations in real time, and EgoSelf~\cite{wang2026egoself} further distills long-term behaviors into profiles for future interaction prediction.

Personalized question answering transforms personal memories into an interactive retrieval interface with natural language. Rather than merely retrieving visually similar content, this task requires smart glasses to identify which past experiences provide relevant evidence for the user's query and how they support the answer. The Ego4D episodic memory benchmark formalizes this problem through introducing queries that require temporal localization of requested information~\cite{grauman2022ego4d}, while EgoLifeQA extends it to long-context daily-life question answering for recalling past events, monitoring habits, and providing personalized assistance~\cite{yang2025egolife}. Compared with conventional retrieval~\cite{karpukhin2020dense,yuan2023semantic}, recent work emphasizes user-centric reasoning: MyEgo~\cite{xiao2026ego} models the wearer as the central actor and distinguishes personal belongings and activities; EgoRetriever~\cite{tang2026memoryaugmented} uses reflective chain-of-thought reasoning to infer user intentions; and WAG~\cite{lu2026query} organizes wearable sensor data into personalized knowledge graphs for answer generation. These studies indicate a shift from memory lookup to personalized contextual reasoning grounded in long-term experience.

Personalized recommendations~\cite{jiang2026atomic,qu2026diffusion,wu2026beyond} extend contextual understanding to proactive suggestions by combining current observations with long-term preference records. Typical examples include recommending menu items aligned with dietary preferences, highlighting news topics of personal interest, or suggesting actions based on historical behavioral patterns. Recent progress in LLMs has further improved user-specific recommendation~\cite{huang2026rerec,qu2025tokenrec,wang2025knowledge,liu2026continuous,ning2026exploring}. For instance, RecICL~\cite{bao2025customizing} supports dynamic adaptation to changing preferences through in-context learning, and CAP-LLM~\cite{wilson2025cap} combines long-term interests with real-world news content for personalized information discovery. Evaluation has also moved toward wearable settings: PerRecBench~\cite{tan2025can} measures preference inference while reducing content-quality and rating biases, VoiceBench~\cite{chen2026voicebench} and Mobile-Bench~\cite{deng2024mobile} establish protocols under different deployment scenarios, and RecBench+~\cite{huang2026towards} introduces realistic interactive environments with optimization strategies.

\subsubsection{Environmental Context}
\label{sec:environmental-context}

Environmental context interprets the user's surroundings as a situated condition rather than a collection of visible objects. Building on spatial perception (\S\ref{sec:spatial-perception}) and conceptual perception (\S\ref{sec:conceptual-perception}), environmental context goes beyond estimating where entities are and identifying what entities or actions are present; it captures how the surrounding scene supports action, persists beyond the current egocentric view, structures object relations and interactions, and signals implicit user needs.

One line of work models environments through functional zones and affordances. EGO-TOPO~\cite{nagarajan2020egotopo} learns topological maps from egocentric video, linking activity-centric zones with likely actions and showing that places can be represented by what they enable the wearer to do. \citet{liu2024grounding} further connects egocentric interaction videos with affordance regions in 3D scenes through the Ego-SAG framework and VSAD dataset. These works suggest that environmental context should encode not only where objects are, but also which actions the surrounding scene makes possible.

A second line of work emphasizes a persistent environmental state. Because an egocentric field of view is narrow and unstable, the current frame may omit relevant parts of the surrounding place. EgoEnv~\cite{nagarajan2023egoenv} addresses this limitation by learning human-centric environment representations that predict potentially unseen local surroundings, while the Aria Everyday Activities dataset~\cite{lv2024aria} illustrates how egocentric video, gaze, speech, trajectories, and 3D scene context provide complementary partial evidence in everyday environments. These works suggest that environmental context should maintain a persistent scene-level state beyond isolated frames.

Environmental context also requires relational and interaction-aware structure, as single action labels often omit how objects participate in activities and which objects jointly define a situation. Egocentric Action Scene Graphs~\cite{rodin2024action} model relationships among actions and interacted objects, while TransFusion~\cite{pasca2024summarize} uses summaries of prior object interactions to represent active-object context. At a finer geometric level, HOT3D~\cite{banerjee2025hot3d} and WHOLE~\cite{ye2026whole} support consistent hand-object relation modeling through egocentric tracking and world-grounded reconstruction. These works suggest that the environmental context should maintain the structural state among hands, objects, and scene elements.

Finally, environmental meaning must be connected with implicit user needs. \citet{sun2025visual} identifies intention-relevant objects while filtering unrelated context and reasoning about uncommon functionality, showing how scene meaning can be grounded in inferred user goals. AiGet~\cite{cai2025aiget} further demonstrates how gaze, environmental context, and user profiles can trigger low-disruption knowledge discovery, while ContextAgent~\cite{yang2025contextagent} and LifeEval~\cite{gao2026lifeeval} evaluate whether sensory context can support proactive decisions and daily-life assistance.

\subsubsection{Temporal Context}
\label{sec:temporal-context}

Temporal context situates the current egocentric observation within a longer continuity of life events. Unlike temporal perception (\S\ref{sec:temporal-perception}), which estimates motion, tracks objects, or forecasts trajectories, temporal context organizes continuous streams into episodes, retrieves relevant past events, maintains compact long-term memory, and identifies recurring patterns for future assistance.

Continuous wearable streams must first be structured into activity episodes. \citet{lv2024aria} records everyday scenarios with Project Aria glasses and synchronizes egocentric video, eye gaze, speech transcripts, and 3D scene context. ActSonic~\cite{mahmud2024actsonic} reaches a similar goal through miniature speakers and microphones on eyeglasses, recognizing activities from inaudible acoustic reflections. Together, these systems show that temporal context is built from multimodal evidence distributed across time, including visual continuity, gaze, speech, trajectories, and acoustic or inertial signals.

Temporal context also requires episodic localization: the system must recover when a relevant event occurred and how it connects to the current query. Episodic Memory Question Answering~\cite{datta2022episodic} formalizes this requirement by building spatio-temporal scene memory and localizing answers within a home tour. SpotEM~\cite{ramakrishnan2023spotem} addresses the efficiency problem of hours-long memory through query-conditioned clip selection and low-cost semantic indexing. A temporal-context layer must therefore support recall while respecting battery, storage, and latency limits.

Beyond individual episodes, the temporal context must maintain a compact long-horizon state. AMEGO~\cite{goletto2024amego} constructs active memories from long egocentric videos by retaining key locations and object interactions, and evaluates sequencing, concurrency, and temporal grounding over the Active Memories Benchmark. ESOM~\cite{manigrasso2026online} extends this direction to online episodic memory by processing each frame once, tracking objects, and storing compact spatio-temporal information for later visual queries. These works emphasize incremental memory updates, selective retention, temporal ordering, and compact interfaces for downstream agents.

Repeated episodes further reveal behavior-level patterns. A useful temporal memory should distinguish isolated events from habits, completed tasks from unfinished ones, and stable routines from recent changes. EgoMemReason~\cite{wang2026egomemreason} evaluates week-long egocentric understanding through entity, event, and behavior memory, where evidence may be sparse and separated by long intervals. Such behavior-level memory helps the system use recurring patterns and recent deviations as contextual cues for later assistance.

\subsubsection{Knowledge Context}
\label{sec:knowledge-context}
Knowledge context gives smart glasses access to evidence beyond the current sensor frame. We organize it into two forms: \emph{external knowledge}, which retrieves information from documents, web pages, manuals, and connected tools; and \emph{internal memory}, which stores retrievable knowledge accumulated from the wearer's longitudinal experiences. Unlike personalized memory (\S\ref{sec:personal-context}), which models user preferences, habits, and profiles, internal memory emphasizes past observations, events, and interactions as evidence for current understanding.

External knowledge is commonly grounded through retrieval-augmented generation (RAG). Classical RAG systems retrieve evidence from non-parametric corpora and condition generation on that evidence~\cite{guu2020realm,izacard2021leveraging,liu2026inference}, while recent variants improve scale, adaptivity, hierarchical retrieval, graph-structured knowledge, and long-context reading~\cite{borgeaud2022improving,asai2023selfrag,edge2024local,yuan2026mkg,jiang2024longrag}. In smart glasses, retrieval must first translate the egocentric frame, gesture, speech, or dialogue into a query. GazePointAR~\cite{lee2024gazepointar} illustrates this grounding step by resolving pronouns and object references from gaze, pointing, and conversation history in wearable AR.

Recent systems further integrate external knowledge into the assistant loop. CRAG-MM~\cite{wang2025cragmm} and WearVox~\cite{lin2025wearvox} introduce multimodal, multi-turn, and multi-hop benchmarks for evaluating retrieval in smart-glasses scenarios. QA-Dragon~\cite{jiang2025qa} and SUPERGLASSES~\cite{jiang2026superglasses} address this problem through query decomposition, object grounding, multimodal web retrieval, and reasoning over retrieved evidence. AiGet~\cite{cai2025aiget} shows a more proactive setting, where gaze and environmental context trigger knowledge discovery before a fully specified query is issued. Egocentric Co-Pilot~\cite{yang2026egocentric} and VisionClaw~\cite{liu2026visionclaw} further connect egocentric perception with web-native tools and downstream actions. The core challenge is therefore perceptually grounded retrieval: resolving what the wearer refers to, selecting the right source, and returning evidence without overloading attention.

Internal memory provides the second form of knowledge context by turning longitudinal egocentric experience into retrievable evidence. General memory-augmented agents store and reuse observations, reflections, and interaction histories across sessions~\cite{park2023generative,shinn2023reflexion,zhong2024memorybank,wang2023augmenting,packer2023memgpt}, while recent systems emphasize structured, evolving, and scalable memory for long-running agents~\cite{xu2025amem,chhikara2025mem0}. In wearable settings, MemX~\cite{chang2021memx} captures attention-relevant moments as compact visual memory under sensing and energy constraints, and EgoLife~\cite{yang2025egolife} introduces EgoRAG for question answering over daily egocentric recordings.

Internal memory also changes interaction. SpeechLess~\cite{kim2026speechless} binds prior interactions to spatial, temporal, activity, and referent context so that users can issue micro-utterances or under-specified requests in repeated situations. AI4Service~\cite{wen2025service} similarly uses a memory unit to support proactive service assistance. These systems suggest a memory pipeline for smart glasses: capture salient events, summarize and index them, retrieve relevant evidence on demand or opportunity, and update or forget memories as context changes.

\textbf{Summary.}
Taken together, these four forms of context shift smart glasses from recognizing isolated egocentric observations to interpreting situated meaning across the wearer, environment, time, and knowledge sources. The central challenge is selective contextualization: preserving enough context for continuity and proactive support while limiting redundancy, retrieval noise, attention burden, and inappropriate intervention.

\subsection{Agentic Intelligence}
\label{sec:agentic-intelligence}

\begin{figure}[t]
    \centering
    \includegraphics[width=\textwidth]{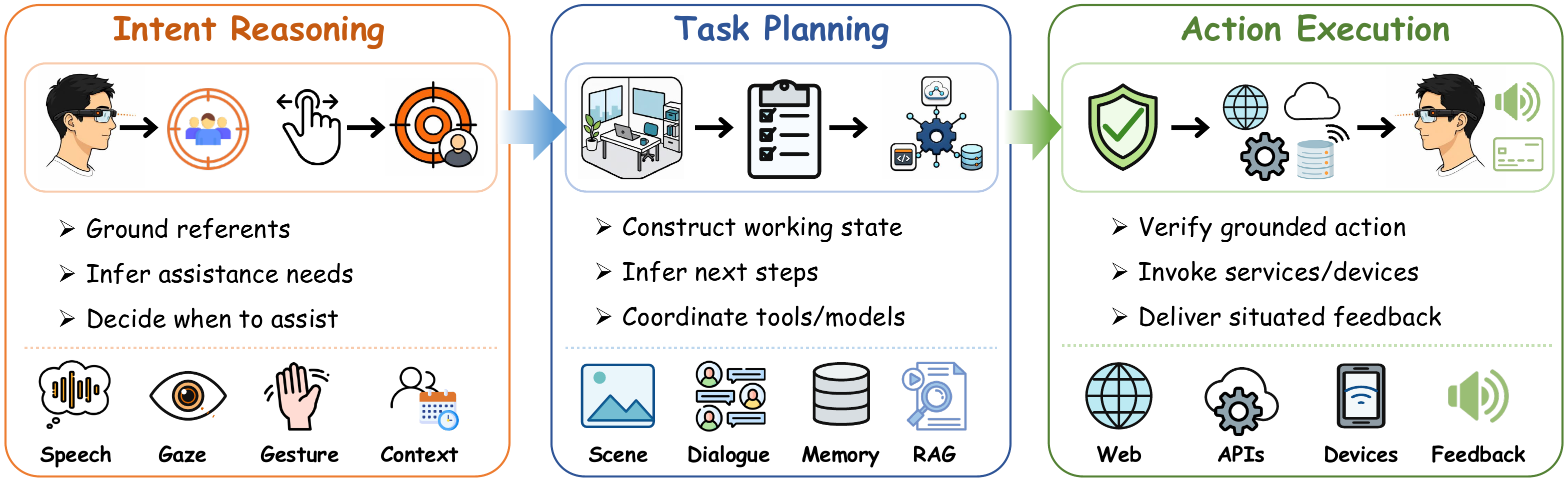}
    \caption{Overview of agentic intelligence in AI smart glasses. Situated observations and contextual evidence are transformed into an intent--plan--act loop: intent reasoning grounds referents and assistance needs, task planning constructs working states and infers next steps under wearable constraints, and action execution invokes services or devices and delivers situated feedback.}
    \label{fig:agentic-intelligence}
\end{figure}

Building on perception and contextual intelligence, agentic intelligence enables smart glasses to move from understanding egocentric situations to pursuing user-centered goals. As illustrated in Figure~\ref{fig:agentic-intelligence}, we organize this capability around an intent--plan--act loop: \emph{intent reasoning} (\S\ref{sec:intent-reasoning}) infers the user's goal and assistance boundary, \emph{task planning} (\S\ref{sec:task-planning}) converts that goal into executable steps under wearable constraints, and \emph{action execution} (\S\ref{sec:action-execution}) carries out those steps through tools, services, connected devices, or situated guidance.

\subsubsection{Intent Reasoning}
\label{sec:intent-reasoning}
Intent reasoning determines what the user is trying to accomplish and whether assistance is appropriate. For smart glasses, intent is often implicit, brief, and grounded in the user's immediate context rather than expressed as a complete instruction. The system must therefore integrate dialogue history, ongoing activities, environmental cues, and personal context to distinguish an explicit command from a passing observation, an expression of curiosity, or a situation where intervention would be unwelcome. This differs from intent recognition in desktop assistants because the relevant evidence may be distributed across what the wearer sees, what the wearer is manipulating, where attention is directed, and what has happened moments or days earlier.

General language-agent work provides the basic abstraction. ReAct~\cite{yao2023react} interleaves reasoning and acting so that an agent can update its beliefs from observations, while embodied-agent research argues that physical agents require world models and user models beyond text-level plans~\cite{fung2025embodied}. In smart glasses, this abstraction first appears as a grounded reference. GazePointAR~\cite{lee2024gazepointar} resolves ambiguous pronouns by combining gaze, pointing, and dialogue context; Gazeify Then Voiceify~\cite{zhang2026gazeify} uses gaze to select a physical object and voice to confirm or describe it; and Persistent Assistant~\cite{cho2025persistent} studies gaze- and gesture-based intent grounding for repeated everyday decisions. These systems suggest that smart glasses must first stabilize the object or event under discussion before passing intent to retrieval, planning, or execution.

A second line of work extends intent reasoning from responding to explicit commands to inferring when assistance should be initiated. ContextAgent~\cite{yang2025contextagent} and WAGIBench~\cite{veerabadran2025benchmarking} evaluate whether wearable agents can identify service needs or high-level user goals from sensory, personal, and digital context. AI4Service~\cite{wen2025service} makes this setting smart-glasses-specific by detecting service opportunities from contextual and memory cues. Together, these systems shift intent reasoning from recognizing what the user asked to deciding whether a useful and acceptable action should be taken.
Intent can also guide system decisions: \citet{jiang2026intention} treats inferred task intent as a variable that determines which visual semantics should be preserved before offloading to a server-side model. Thus, intent reasoning acts as the entry point of the agentic loop by grounding what the user means, estimating whether assistance should occur, and passing a confidence-aware goal representation to downstream planning.

\subsubsection{Task Planning}
\label{sec:task-planning}
Task planning translates grounded intent into executable steps. General agent frameworks provide useful foundations for goal decomposition, model coordination, and tool invocation~\cite{jiang2025hibench,shen2023hugginggpt,schick2023toolformer}. For smart glasses, however, planning must coordinate perception, reasoning, interaction, and device constraints, turning situated goals into compact executable workflows.

The distinctive planning problem is working state construction. Unlike desktop agents that operate over relatively stable interface states, glasses agents plan from partial and changing egocentric evidence. The system must construct a working state based on perceptual evidence, task progress, and relevant memories before selecting the next step. EPD~\cite{shi2024epd} extracts task-relevant memory from progress videos for context-aware egocentric planning. CAPEAM~\cite{kim2023context} emphasizes explicit task context and environment-aware memory of spatial and object-state changes. MEMORA~\cite{yu2026memora} extends this line by retrieving typed embodied action memories to support reasoning and planning.

After constructing the working state, the agent must infer feasible next steps and anticipate how the task may unfold. VLaMP~\cite{patel2023pretrained} formulates visual planning for assistance as generating future action sequences from a natural-language goal and a video of the user's progress. Intention-conditioned anticipation~\cite{mascaro2023intention} uses high-level intent to produce more coherent long-term action sequences, while NAOGAT~\cite{thakur2024leveraging} predicts next-active objects and future interactions from egocentric video. Planning benchmarks such as EgoPlan-Bench~\cite{chen2026egoplan}, EgoPlan-Bench2~\cite{qiu2026egoplan}, and LifeEval~\cite{gao2026lifeeval} further evaluate this ability at the level of task progression, goal completion, and recovery over time.

Finally, task planning must coordinate the models, tools, and interaction steps needed to execute the plan. SUPERGLASSES~\cite{jiang2026superglasses} decomposes visual questions into object identification, multimodal web search, and evidence-based reasoning. Egocentric Co-Pilot~\cite{yang2026egocentric} coordinates long-horizon egocentric context with perception, reasoning, and web tools, while Vinci~\cite{huang2024vinci} illustrates real-time planning under short interaction cycles. These systems suggest that smart-glasses planning should output a compact plan over state, next steps, tool or model choices, and possible clarification points, leaving concrete service calls and feedback delivery to action execution.

\subsubsection{Action Execution}
\label{sec:action-execution}
Action execution closes the agent loop by carrying out the planned step through digital tools, connected devices, or user-facing guidance. General agent research focuses on reliable tool selection, command generation, and environment interaction, enabling agents to perform web search and navigation~\cite{nakano2021webgpt}, API use~\cite{schick2023toolformer,patil2023gorilla}, and interface actions~\cite{deng2023mind2web,zhou2023webarena}, with recent work further improving execution reliability and efficiency~\cite{chen2026tumix,wu2026dynamic,wu2026timing,ning2026efficiency}. Smart glasses add a harder grounding problem: the action target may be a physical object, spoken request, sensitive scene, or remembered context rather than a clean browser state.

Execution in smart glasses, therefore, begins before the tool call. The agent must ensure that the perceived target and planned operation are grounded in the right situation. WearVQA~\cite{chang2026wearvqa} shows how wearable image degradations can affect downstream visual decisions, while WearVox~\cite{lin2025wearvox} evaluates tool calling and search-grounded question answering under egocentric multichannel audio. MICA~\cite{sarch2025grounding} further shows that gaze and speech cues improve task assistance by grounding demonstrations in intent and contextual evidence. These works suggest that execution reliability depends on correctly identifying the object, request, and situation that triggered the action.

Recent smart-glasses agents connect this grounded state to concrete services. SUPERGLASSES~\cite{jiang2026superglasses} couples object detection, query decomposition, multimodal web search, and evidence-based reasoning. Ego2Web~\cite{yu2026ego2web} pairs egocentric video with online workflows, and Egocentric Co-Pilot~\cite{yang2026egocentric} connects long-horizon egocentric context to web-mediated assistance. VisionClaw~\cite{liu2026visionclaw} further supports shopping, document note-taking, meeting assistance, event creation, and device control through live egocentric perception and external agents. These systems show execution for AI smart glasses as a controlled handoff from situated perception to external action.

Execution also includes user-facing guidance when the planned step cannot be completed through tools or devices alone. Satori~\cite{li2025satori} generates proactive step-by-step AR guidance from user actions, task goals, objects, and environmental context. Eye2Eye~\cite{teng2026seeing} treats situated feedback as part of first-person cognitive alignment, where multimodal cues help maintain shared attention and common ground. For smart glasses, guidance should remain tied to the perceived object, location, or activity state, so that instructions, warnings, or confirmations are concise and safe to act on.

\textbf{Summary.}
Agentic intelligence closes the loop from intent to plan to action. Its core challenge is calibrated autonomy: smart glasses must ground user intent, construct compact plans, and execute situated actions while avoiding distraction, privacy exposure, inappropriate interruption, and unsafe behavior under uncertain perception and context.

\section{Interaction Design}
\label{sec:interaction-design}

\begin{figure}[t]
    \centering
    \includegraphics[width=\textwidth]{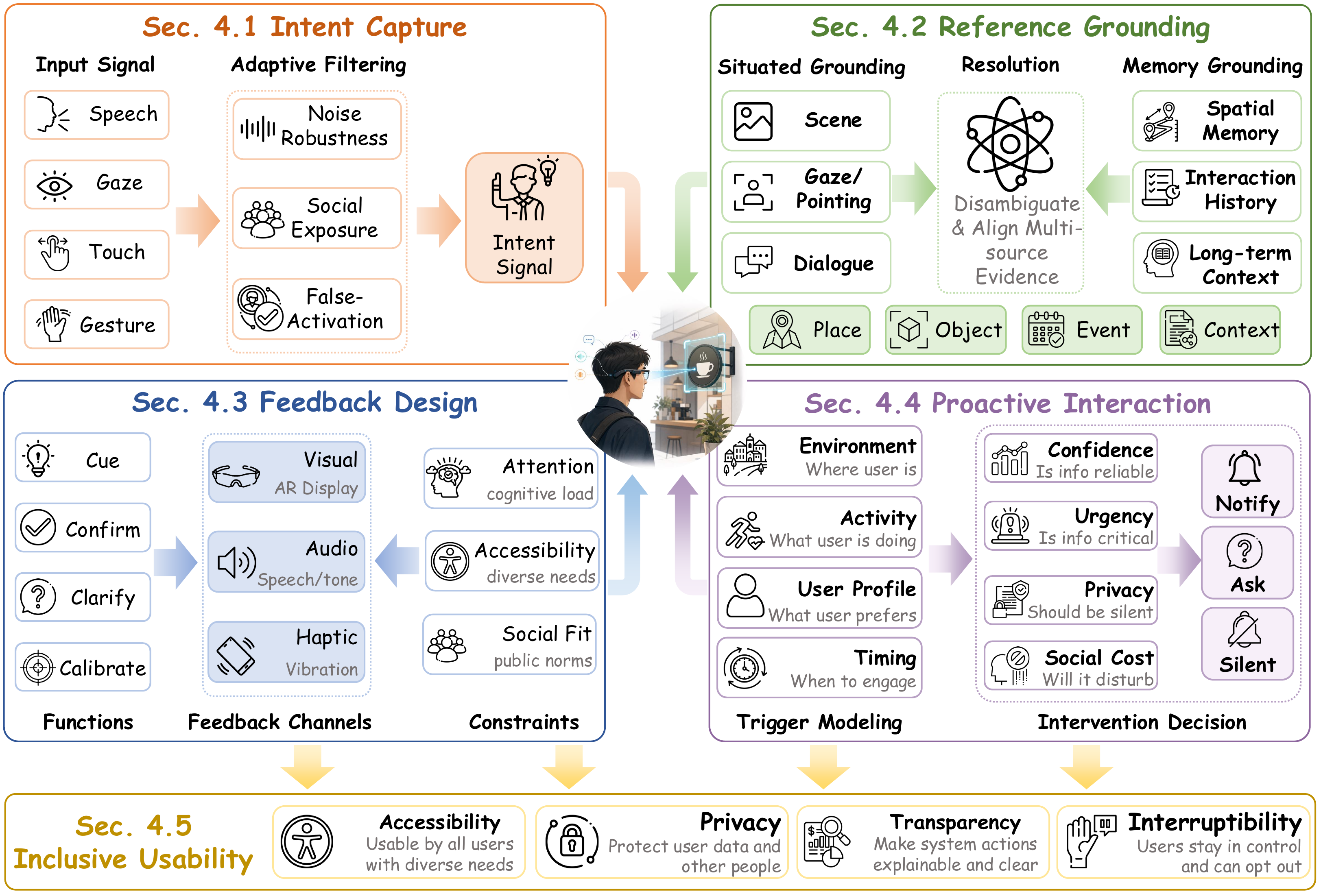}
    \caption{Overview of interaction design in AI smart glasses. The interaction loop captures intent, grounds references, selects feedback, regulates proactive intervention, and keeps assistance accessible, privacy-aware, transparent, and interruptible in wearable settings.}
    \label{fig:interaction-design}
\end{figure}

Interaction design is the user-facing control layer through which AI smart glasses capture intent, ground references, deliver feedback, regulate proactive intervention, and remain usable in daily settings. Unlike phones or desktop assistants, smart glasses operate during mobile, social, and attention-constrained activities, where interaction may be brief, multimodal, implicit, or constrained by public acceptability.
As shown in Figure~\ref{fig:interaction-design}, we organize interaction design into five connected functions. \emph{Intent Capture}(\S\ref{sec:intent-capture}) detects engagement through speech, gaze, touch, hand gestures, and other cues. \emph{Reference Grounding}(\S\ref{sec:reference-grounding}) resolves intended referents through situated evidence and memory. \emph{Feedback Design}(\S\ref{sec:feedback-design}) combines channel selection with repair-oriented feedback for confirmation, correction, and trust calibration. \emph{Proactive Interaction}(\S\ref{sec:proactive-interaction}) couples trigger modeling with intervention decisions. \emph{Inclusive Usability}(\S\ref{sec:inclusive-usability}) treats user diversity, accessibility, privacy, transparency, and control as cross-cutting constraints. Earlier surveys mainly characterized smart-glasses interaction as micro-interactions and device control~\cite{lee2018interaction}; recent AI-mediated systems instead treat interaction as a closed loop among multimodal sensing, contextual reasoning, user correction, and situated assistance.

\subsection{Intent Capture}
\label{sec:intent-capture}

Intent capture concerns how smart glasses infer that the wearer wants to engage the system and what interaction form that intention should take. Speech is a natural starting point because it supports hands-free control, but it is vulnerable to noise, socially exposed in public, and ambiguous when nearby conversations occur. WearVox~\cite{lin2025wearvox} captures this difficulty by evaluating wearable voice-assistant tasks such as search-grounded question answering, tool calling, side-talk rejection, and speech translation. EchoSpeech~\cite{zhang2023echospeech} complements audible speech with eyewear-mounted acoustic sensing for silent speech recognition, reducing dependence on vocalized commands in public or noisy environments. These works show that speech-based intent capture must address not only recognition accuracy, but also addressability and social exposure.

Gaze provides a second intent channel because smart glasses are naturally aligned with the wearer's visual attention. GazeTrak~\cite{li2024gazetrak} explores acoustic gaze estimation on a glasses frame, while EyeGesener~\cite{sun2024eyegesener} targets intentional eye gestures and filters daily eye movements to reduce false activation. Together, these systems show how gaze and eye motion can support low-friction intent capture, while also raising calibration and reliability challenges.

Touch and hand gestures provide more deliberate forms of control and confirmation. FingerGlass~\cite{xu2025fingerglass} uses a frame-mounted fingerprint sensor for finger-aware touch interaction, while Helios~\cite{bhattacharyya2025helios} studies low-power gesture recognition for always-on smart eyewear. GlassMessaging~\cite{janaka2023glassmessaging} further shows that everyday smart-glasses interaction often combines voice and manual input during mobile activities. Together, these systems suggest that intent capture should coordinate input channels adaptively rather than treat them as independent choices.

\subsection{Reference Grounding}
\label{sec:reference-grounding}

Reference grounding connects a captured intention to the place, object, event, or prior experience that the wearer means. This is a core smart-glasses problem because wearable commands are often short and deictic. Grounding spans both the current scene and prior experience: situated grounding resolves references in the immediate environment, whereas memory grounding links an utterance or gesture to earlier interaction history. Without such grounding, accurate speech recognition still produces an incomplete query because the system cannot identify what should be passed to perception, reasoning, or agentic planning.

Situated grounding addresses references to visible objects and ongoing activities. GazePointAR~\cite{lee2024gazepointar} demonstrates this requirement by combining gaze, pointing, voice, and conversation history to resolve ambiguous pronouns in wearable augmented reality. Its significance lies not only in multimodal fusion but also in showing that reference resolution is a prerequisite for assistance in physical space. Gazeify Then Voiceify~\cite{zhang2026gazeify} extends this principle to displayless smart glasses, where gaze selects a physical object, and voice confirms or describes it without visual overlays. This design is important for audio-first glasses, where the system cannot rely on displaying a highlighted target and must instead support spoken confirmation and repair.

Memory grounding expands reference resolution beyond the current frame. SpeechLess~\cite{kim2026speechless} uses personalized spatial memory to support micro-utterances in everyday augmented reality, reducing the need for long public commands when repeated places, objects, and activities are already known. Persistent Assistant~\cite{cho2025persistent} similarly treats grounding as an ongoing process supported by embodied input and multimodal feedback. These systems suggest that reference grounding in AI smart glasses should align situated evidence with longer-term memory, rather than treating each utterance as a one-shot mapping from language to command.

\subsection{Feedback Design}
\label{sec:feedback-design}

Feedback design determines how smart glasses return information, request clarification, and expose uncertainty without overloading the wearer. For this form factor, the central question is how to select feedback channels and support repair, confirmation, and trust calibration without pulling attention away from the surrounding world.

Feedback channel selection is constrained by the wearable form factor. Visual overlays can support spatial guidance but may occlude the scene; audio can preserve displayless use but may interfere with conversations or environmental awareness; haptics and subtle notifications can reduce visual load but carry limited information. Accessibility-oriented navigation illustrates this trade-off: visual and audio wayfinding guidance on smart glasses for people with low vision affects not only navigation performance, but also attention, confidence, and the coordination between assistance and residual vision~\cite{zhao2020wayfinding}. GlassMessaging~\cite{janaka2023glassmessaging} reaches a related conclusion in everyday communication, where head-mounted messaging can improve access during multitasking, but must balance speed, accuracy, visual demand, and the appropriateness of voice input.

Repair-oriented feedback becomes necessary when AI assistance is grounded, uncertain, or awaiting user confirmation. Persistent Assistant~\cite{cho2025persistent} treats multimodal feedback as part of the intent-grounding loop rather than a final display stage. Gazeify Then Voiceify~\cite{zhang2026gazeify} shows the same principle in displayless interaction, where gaze-grounded selection is converted into voice-mediated confirmation. Feedback should therefore act as a lightweight repair channel, providing enough evidence for correction and trust calibration without competing with ongoing activity.

\subsection{Proactive Interaction}
\label{sec:proactive-interaction}

Proactive interaction occurs when smart glasses initiate, compress, or reshape assistance from context rather than waiting for a fully specified command. Its design depends on trigger modeling, which detects when assistance may be relevant, and intervention decision, which determines whether and how the system should act. Proactivity can reduce interaction effort, but poorly timed prompts can distract the wearer, expose private context, or make the device socially intrusive.

Trigger modeling uses situated and personal context to identify moments where assistance may be useful. AiGet~\cite{cai2025aiget} exemplifies this direction by using gaze, environmental context, and user profiles to surface knowledge during everyday moments. SpeechLess~\cite{kim2026speechless} approaches the same problem through spatial memory, allowing short utterances to trigger contextually appropriate assistance. In both cases, context reduces interaction effort only when the system can infer relevance from the wearer's activity, surroundings, and prior experience.

Intervention decision governs whether a detected opportunity should become an action, a clarification request, or no interruption at all. AI4Service~\cite{wen2025service} uses context and memory to provide proactive service assistance with AI smart glasses, while Persistent Assistant~\cite{cho2025persistent} studies persistent everyday assistance through intent grounding and feedback. In this section, these systems are best understood as interaction designs for timing, interruption management, confirmation, and user control. Future systems should condition proactive behavior on confidence, privacy sensitivity, and social context, deciding when to remain silent, ask for confirmation, or intervene.

\begin{table*}[t]
\centering
\caption{Representative systems organized by interaction-design dimensions for AI smart glasses.}
\label{tab:representative-interaction}
\renewcommand{\arraystretch}{1.15}
\scriptsize
\setlength{\tabcolsep}{3pt}
\begin{tabularx}{\textwidth}{>{\centering\arraybackslash}p{0.15\textwidth} >{\centering\arraybackslash}p{0.20\textwidth} >{\centering\arraybackslash}X >{\centering\arraybackslash}p{0.22\textwidth}}
\toprule
\textbf{\makecell{Interaction\\dimension}} & \textbf{Design focus} & \textbf{Representative systems} & \textbf{Main constraints} \\
\midrule
\multirow{3}{=}{Intent capture}
& Speech input & WearVox~\cite{lin2025wearvox}; EchoSpeech~\cite{zhang2023echospeech} & Noise, addressability, exposure \\
& Gaze input & GazeTrak~\cite{li2024gazetrak}; EyeGesener~\cite{sun2024eyegesener} & Calibration, false activation \\
& Touch and gesture input & FingerGlass~\cite{xu2025fingerglass}; Helios~\cite{bhattacharyya2025helios} & Power, learnability, fatigue \\
\midrule
\multirow{2}{=}{Reference grounding}
& Situated grounding & GazePointAR~\cite{lee2024gazepointar}; Gazeify Then Voiceify~\cite{zhang2026gazeify} & Deixis, confirmation \\
& Memory grounding & SpeechLess~\cite{kim2026speechless}; Persistent Assistant~\cite{cho2025persistent} & Personalization, repair \\
\midrule
\multirow{2}{=}{Feedback design}
& Channel selection & Zhao et al.~\cite{zhao2020wayfinding}; GlassMessaging~\cite{janaka2023glassmessaging} & Attention load, accessibility \\
& Repair-oriented feedback & Persistent Assistant~\cite{cho2025persistent}; Gazeify Then Voiceify~\cite{zhang2026gazeify} & Confirmation, trust calibration \\
\midrule
\multirow{2}{=}{Proactive interaction}
& Trigger modeling & AiGet~\cite{cai2025aiget}; SpeechLess~\cite{kim2026speechless} & Timing, context reliability \\
& Intervention decision & AI4Service~\cite{wen2025service}; Persistent Assistant~\cite{cho2025persistent} & Interruption, privacy, control \\
\midrule
\multirow{2}{=}{Inclusive usability}
& User diversity and accessibility & Zhao et al.~\cite{zhao2020wayfinding}; Guo et al.~\cite{guo2026olderworkers} & Cognitive load, comfort \\
& Privacy, transparency, and control & Lens Privacy~\cite{zhang2025lensprivacy} & Consent, legibility, social acceptability \\
\bottomrule
\end{tabularx}
\vspace{-1mm}
\end{table*}

\subsection{Inclusive Usability}
\label{sec:inclusive-usability}

Inclusive usability evaluates whether interaction remains practical across users, abilities, environments, and social settings. Smart glasses are worn on the face, near the eyes and ears, and often in the presence of bystanders, so interaction choices have consequences beyond task efficiency. The main constraints are user diversity and accessibility, which determine whether input and feedback remain usable across abilities and contexts, together with privacy, transparency, and control, which determine whether sensing and AI assistance remain understandable and interruptible.

User diversity and accessibility require interaction techniques to be evaluated against real user capabilities rather than aggregate performance alone. Low-vision wayfinding demonstrates that smart-glasses guidance should account for sensory ability, navigation context, attention, and confidence~\cite{zhao2020wayfinding}. Interaction design work for AI smart glasses and older workers similarly emphasizes embodied cognition, cognitive load, physical comfort, and socio-emotional needs, showing that efficient assistance can still fail when age-related usability requirements are ignored~\cite{guo2026olderworkers}. Input systems such as EchoSpeech~\cite{zhang2023echospeech} and FingerGlass~\cite{xu2025fingerglass} can reduce dependence on public speech or large gestures, but they also introduce calibration, learnability, and fatigue constraints.

Privacy, transparency, and control cut across all modalities. Vision-language model interactions on smart glasses can expose bystanders, sensitive documents, or private scenes, especially when image capture and cloud processing are hidden behind a simple voice query~\cite{zhang2025lensprivacy}. This makes consent, legibility, and user control central to interaction design. Interfaces should help wearers and nearby people understand when sensing is active, what information is used, and how assistance can be paused, corrected, deleted, or bypassed.

\textbf{Summary.}
Taken together, these five functions define interaction design as the coordination of input, grounding, feedback, proactive intervention, and inclusive usability under wearable constraints. The central challenge is to keep this loop usable during continuous real-world activity, where assistance must remain easy to address, correctly grounded, repairable, well-timed, accessible, privacy-aware, and socially acceptable. Table~\ref{tab:representative-interaction} summarizes representative systems across this interaction loop.
\section{Application Scenarios}
\label{sec:applications}

\begin{figure}[t]
    \centering
    \includegraphics[width=\textwidth]{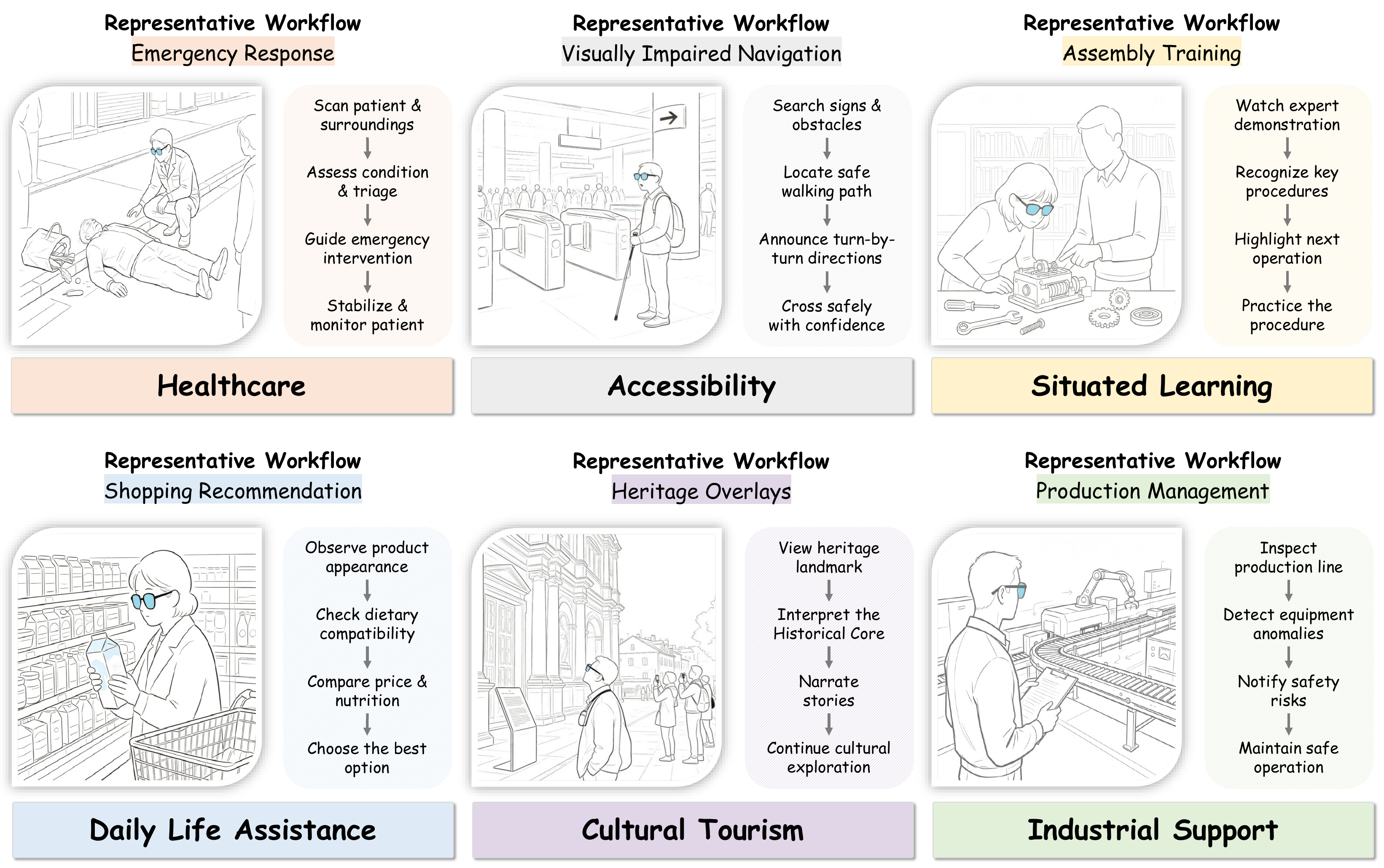}
    \caption{Representative workflow of AI smart glasses across healthcare, accessibility, situated learning, daily life assistance, cultural tourism, and industrial support.}
    \label{fig:application_scenarios}
\end{figure}

Application scenarios evaluate whether smart glasses can turn egocentric sensing into timely, situated assistance across healthcare, accessibility, learning, daily life, tourism, and industry. Figure~\ref{fig:application_scenarios} illustrates a representative workflow across these domains. The following subsections examine use scenarios, representative systems, and deployment concerns to show how wearable intelligence and interaction design must adapt to domain-specific constraints.

\subsection{Healthcare}
\label{sec:healthcare-applications}

Healthcare is a demanding domain for AI smart glasses, where assistance must remain hands-free, privacy-preserving, and responsive to time-critical decisions. By preserving the clinician's egocentric view, smart glasses can deliver patient information, procedural guidance, remote expertise, and alerts at the point of care. Existing studies explore care delivery, monitoring, telemedicine, and personalized intervention, but deployment remains limited by privacy, validation, standardization, battery life, and medical ethics~\cite{wang2025digitalhealth}.

Emergency care illustrates this requirement. At an incident scene, an emergency medical technician must treat the patient while forming a rapid account of symptoms, measurements, bystander reports, and triage requirements. EMSGlass~\cite{jin2025emsglass} frames this setting as a multimodal smart-glasses system: EMSNet integrates scene evidence, text, and vital signs to infer the incident state, while EMSServe enables low-latency field deployment. The workflow translates clinical and environmental cues into actionable guidance while keeping feedback concise, interruptible, and urgency-aware.
Other healthcare applications broaden the setting while retaining the need for situated assistance. In nursing education, smart glasses provide egocentric views of clinical procedures for demonstration, remote supervision, and reflection~\cite{romare2023nursing}. Adoption studies further show that acceptance depends partly on how the device affects professional–patient interaction~\cite{zuidhof2025acceptance}. For older adults and long-term care, scoping reviews suggest potential benefits for independence and daily functioning, though evidence remains fragmented~\cite{burch2025cognitive,delgadomorales2026active}. These applications also highlight temporal context, as assistance may depend on routines, medication histories, and behavioral deviations.

Healthcare shows both opportunity and caution. AI smart glasses can provide hands-free, context-aware support when clinicians, caregivers, or patients cannot shift attention to another device, but errors in perception, timing, command interpretation, or data handling can directly affect care quality and trust. Clinical deployment, therefore, requires validated perception and reasoning, auditable interaction histories, conservative proactivity, and feedback designs that preserve professional control while protecting patient privacy under realistic workflow pressure.

\subsection{Accessibility}
\label{sec:accessibility-applications}

Accessibility is a natural domain for smart glasses to serve as assistive infrastructure rather than convenience devices. For users with visual, language, motor, or cognitive impairments, their key advantage is situated access: observing the wearer's environment and providing guidance without requiring a handheld device. Prior low-vision wayfinding studies show that smart glasses can reshape attention and combine residual vision with device feedback~\cite{zhao2020wayfinding}. AI further extends this role toward situated interpretation, question answering, and social participation.

LLM-based assistance for visually impaired users illustrates this workflow. \citet{lee2025llmvi} shows how smart glasses can connect egocentric sensing with language-model assistance, enabling users to ask situated questions about nearby objects, text, routes, or environmental cues. Such systems must ground the user's query in the current scene and return brief, confidence-aware responses through audio or sparse visual feedback. Broader accessibility work shows why model capability alone is insufficient: co-design with people with aphasia reveals barriers such as gesture burden, visual clutter, form factor, and social conspicuousness~\cite{curtis2025accessible}, while CollabLens shows how smart glasses can reshape collaboration between blind or low-vision users and sighted peers~\cite{ding2026mixedvision}. Accessibility, therefore, depends not only on individual task completion but also on social participation, interaction burden, and inclusive design.

Assistive smart glasses should be judged by whether they increase independence and agency without adding new burdens. Reliable perception matters, but it must be paired with interaction designs that remain private, socially legible, repairable, and adaptable across users. The main risks are overconfident guidance, excessive interruption, inaccessible feedback, and assumptions that one modality or assistance style works for everyone.

\subsection{Situated Learning}
\label{sec:situated-learning-applications}

Situated learning naturally fits smart glasses because they share the learner's egocentric view and can attach guidance directly to the object, action, or concept being studied. Reviews of AR smart glasses in industrial assembly demonstrate their value for procedural learning, while highlighting challenges in authoring, tracking, feedback readability, and production integration~\cite{danielsson2020industrialassembly}. AI smart glasses extend static overlays toward adaptive instruction that tracks attention, task state, and relevant knowledge.

Representative work illustrates how smart glasses can support procedural and conceptual learning rather than only task execution. MARMA~\cite{konstantinidis2020marma} aligns recognized equipment with step-by-step repair guidance, showing how procedural knowledge can be anchored to visible objects. Mixed-reality learning attaches abstract concepts to visible procedures and spatial demonstrations~\cite{tang2020mrlearning}, while smart-classroom research emphasizes pedagogy, teacher control, and institutional readiness~\cite{dimitriadou2023smartclassroom}. Across classrooms, laboratories, and training settings, the central requirement is to keep guidance pedagogically useful: feedback should explain, confirm, or correct learner understanding without simply optimizing task speed. Adoption studies reinforce this distinction, showing that task fit, workload, comfort, and social setting shape whether learning support is accepted~\cite{koutromanos2023acceptance,guo2026olderworkers}.

AI smart glasses are most valuable for situated learning when they operate as contextual tutors rather than passive displays. Their effectiveness depends on whether guidance is synchronized with task state, learner attention, and pedagogical goals, while keeping errors, cognitive burden, interruption cost, and knowledge-base maintenance under control.

\subsection{Daily Life Assistance}
\label{sec:everyday-assistance-applications}

Daily life assistance is the broadest application area for AI smart glasses, arising from opportunistic needs during shopping, commuting, reading, and social interaction. Because glasses are aligned with the user's gaze, speech, movement, and surroundings, the key challenge is to determine what information is relevant, when to intervene, and how to use personal or external knowledge appropriately.

Proactive knowledge discovery during shopping or commuting exemplifies this role. AiGet~\cite{cai2025aiget} turns everyday moments into contextual knowledge opportunities: when users glance at a product, poster, or unfamiliar place, the system selectively surfaces concise suggestions based on expected utility and interruption cost. Users can then refine or expand the information through brief voice interactions grounded in gaze and dialogue history.
Other work expands daily life assistance from momentary suggestions to longitudinal memory and agentic services. EgoLife~\cite{yang2025egolife} combines long-duration daily-life recordings with EgoGPT and EgoRAG, showing why future glasses need retrieval over personal episodes rather than isolated frames. SUPERGLASSES~\cite{jiang2026superglasses} connects objects in view with external multimodal evidence, while VisionClaw~\cite{liu2026visionclaw} links live perception to downstream tools for daily service tasks. Together, these systems move daily life assistance beyond recognition toward memory, retrieval, planning, and execution.

Daily life assistance highlights the promise and fragility of always-available wearable intelligence. The value lies in timely, personalized support across small everyday moments, but trust can erode quickly when suggestions are mistimed, memory is used inappropriately, or inferred context is difficult to inspect and correct. Useful systems must therefore keep assistance brief, socially acceptable, privacy-aware, and controllable across repeated daily use.

\subsection{Cultural Tourism}
\label{sec:cultural-tourism}

In cultural tourism, smart glasses can transform visits into situated interpretation by linking exhibits, landmarks, routes, and reconstructions to the visitor's view. Their effectiveness depends on useful content, device comfort, and social fit~\cite{tom2016mapping,han2019arsgvisitor}. The key challenge is to infer visitor attention, retrieve reliable cultural knowledge, and provide interpretation at an appropriate level of detail.

Smart-glasses tourism systems illustrate this workflow. TouristicAR~\cite{obeidy2018touristicar} delivered context-aware tourism content at UNESCO World Heritage sites by connecting physical landmarks with digital information. \citet{litvak2020enhancing} similarly used visitor location and viewing orientation to identify points of interest and present explanations. In AI smart glasses, this workflow can extend toward recognizing attended landmarks, retrieving cultural records, supporting follow-up questions, and adapting explanations to visitor intent, language, and movement. Museum guides, heritage-site assistants, and scenic-area services differ in content and interaction needs, but all require the system to align perception, localization, cultural knowledge, narrative generation, and navigation.

Cultural tourism requires AI smart glasses to balance situated interpretation with factual reliability and social fit. The assistant must adapt to visitor attention, movement, language, and interests, while preserving historical authenticity, distinguishing verified knowledge from generated explanation, and remaining robust to localization error, lighting variation, network instability, and crowded public settings.

\subsection{Industrial Support}
\label{sec:industry-support-application}

Industrial smart glasses support operational work by giving workers hands-free access to tasks, equipment, and safety information. Early systems overlaid equipment status and operating guidance, including livestock information in farming~\cite{caria2019exploring,caria2020performance}. The central goal is reliable work execution: the system must understand work context, detect abnormal states, and provide timely guidance without disrupting the task.

Existing industrial systems illustrate different stages of this operational loop. In aquaculture, \citet{xi2023smart} used a smart headset to collect prawn-related observations during feeding-tray inspection and convert them into evidence about pond conditions. In shipbuilding, \citet{kunkera2024development} connected augmented reality with ship models, digital twins, and production workflows so that workers could localize components and update assembly status in context. In mining and construction, \citet{baek2020smart} used smart glasses and Bluetooth beacons to warn workers when equipment proximity became dangerous. Together, these cases show that industrial support must identify work objects, connect them to operational knowledge, and provide time feedback according to workplace risks.

Industrial deployment leaves little room for fragile assistance. Because incorrect recognition, delayed warnings, or inappropriate procedural guidance may affect worker safety and production reliability, AI smart glasses must remain usable under obstruction, drift, lighting variation, noise, network instability, battery limits, comfort constraints, and legacy-system requirements. The key test is whether workers can rely on the system for timely guidance while retaining the ability to inspect, override, or correct its decisions.
\section{Open Problems and Future Directions}
\label{sec:future-direction}

\begin{figure}
    \centering
    \includegraphics[width=\linewidth]{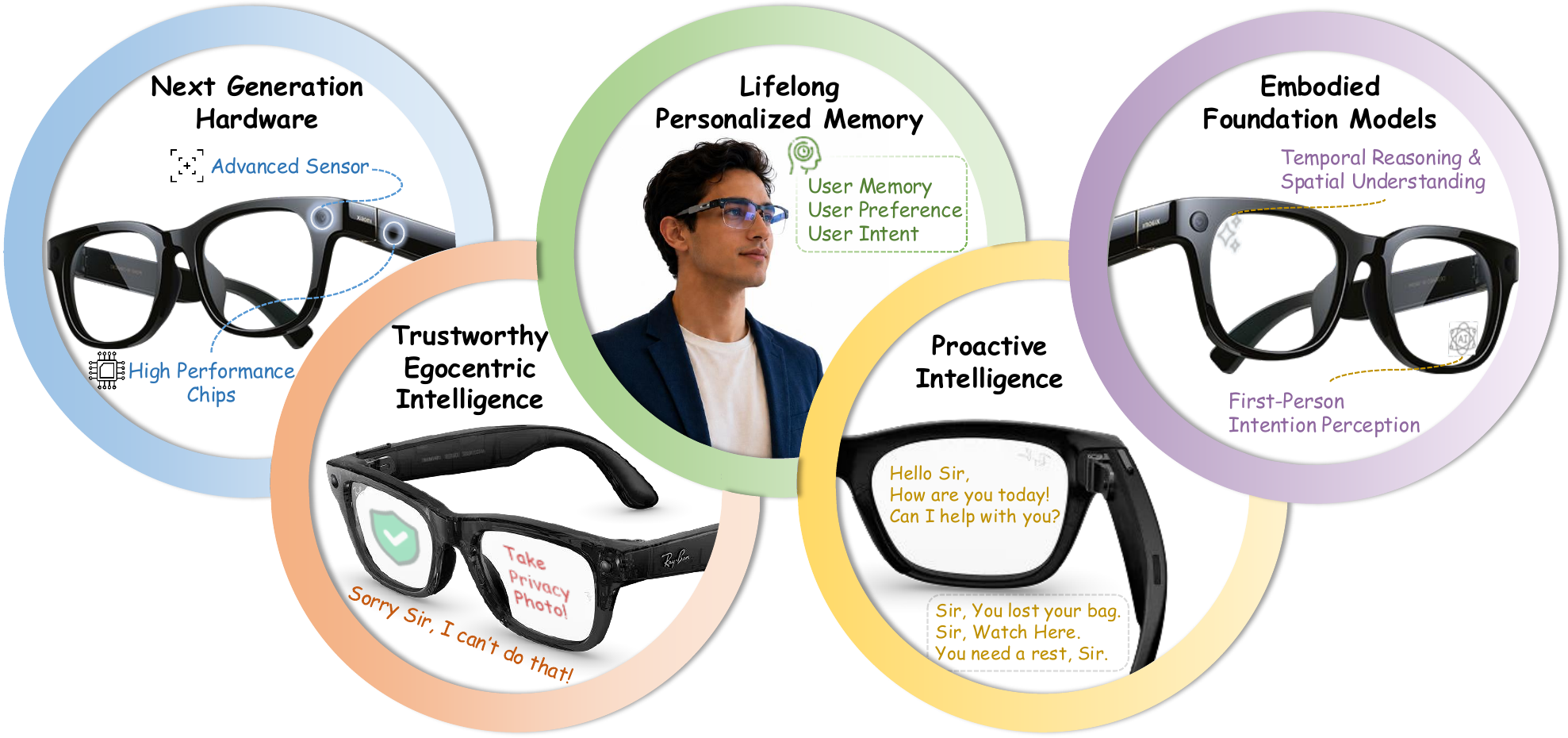}
    \caption{Overview of five cross-cutting open problems for AI smart glasses: next-generation hardware, trustworthy egocentric intelligence, lifelong personalized memory, proactive intelligence, and embodied foundation models.}
    \label{fig:future_direction}
\end{figure}

Building on the preceding sections, this section discusses several promising future research directions for AI smart glasses, as illustrated in Figure~\ref{fig:future_direction}.

\subsection{Next-Generation Hardware}
\label{sec:next-generation-hardware}

\textbf{Status and Challenges.}
As AI smart glasses evolve toward persistent multimodal assistants, their hardware faces a \emph{capability--wearability trade-off}: stronger perception and reasoning require more computation, memory, and sensing resources, while all-day devices must remain lightweight and energy-efficient. Three bottlenecks are central. \textbf{(1)} High-resolution displays and always-on sensors improve interaction but increase power consumption and device complexity. \textbf{(2)} On-device multimodal inference requires stronger processors and storage, yet sustained computation is constrained by battery and thermal budgets. \textbf{(3)} Battery capacity, heat dissipation, connectivity, weight, and form factor are tightly coupled, making hardware design a system-level optimization problem.

\textbf{Outlook.}
\textbf{(1) Hardware specialization} can reduce common workload costs through efficient displays, sensors, neural processors, and vision accelerators. \textbf{(2) Heterogeneous computation} can distribute inference across glasses, phones, edge devices, and the cloud according to latency, privacy, and energy constraints. \textbf{(3) Hardware--software co-design} can combine adaptive sensing, model compression, conditional computation, and resource scheduling. The goal is to balance \emph{intelligence, efficiency, and wearability}, rather than maximize any single hardware specification.

\subsection{Trustworthy Egocentric Intelligence}
\label{sec:trustworthy-ai-glasses}

\textbf{Status and Challenges.}
As AI smart glasses perceive users and surroundings continuously, trustworthiness becomes a first-class concern. Their egocentric and always-on nature creates a tension between \emph{personalized intelligence and responsible operation}. Three challenges are central. \textbf{(1) Fairness:} perception and decision-making should remain reliable across users, environments, and social groups without biased assumptions about gender, age, occupation, or culture. \textbf{(2) Explainability:} users should understand why an object is recognized, a recommendation is made, or an action is suggested, especially in high-stakes scenarios. \textbf{(3) Safety, robustness, and privacy:} smart glasses must remain reliable under noisy inputs, dynamic environments, and adversarial instructions while protecting users and bystanders.

\textbf{Outlook.}
\textbf{(1) Trustworthy-by-design models} should incorporate fairness, uncertainty estimation, and explainability into perception and reasoning pipelines. \textbf{(2) Privacy-aware sensing and computation} can combine selective sensing, on-device processing, data minimization, and fine-grained access control. \textbf{(3) Real-world trustworthiness evaluation} should assess fairness, robustness, safety, privacy, and explainability under diverse users and changing environments. Smart glasses should be accurate, predictable, accountable, and safe for both wearers and bystanders.

\subsection{Lifelong Personalized Memory}
\label{sec:personalized-memory-ai-glasses}

\textbf{Status and Challenges.}
AI smart glasses are naturally suited to long-term personalized assistance because they continuously observe users' environments, behaviors, routines, and preferences. Yet most systems remain limited to short-term interaction contexts, creating a gap between \emph{continuous perception and persistent personalization}. Three challenges are central. \textbf{(1) Memory selection:} the system must determine which experiences are worth retaining. \textbf{(2) Memory evolution:} stored knowledge should be updated, consolidated, compressed, or forgotten as users and environments change. \textbf{(3) Memory retrieval and control:} relevant memories must be recalled efficiently, while users can inspect, correct, and delete what the system remembers.

\textbf{Outlook.}
\textbf{(1) Adaptive memory management} should decide when to store, retrieve, summarize, update, or forget information according to relevance and long-term utility. \textbf{(2) Hierarchical and multimodal memory} can organize episodic experiences, semantic knowledge, user preferences, and multimodal observations at different temporal and semantic scales. \textbf{(3) User-controllable memory} should remain transparent and editable, giving users control over what is retained and how it shapes future assistance. Lifelong memory should accumulate useful experience without sacrificing efficiency, adaptability, or user agency.

\subsection{Proactive Intelligence}
\label{sec:proactive-intelligence}

\textbf{Status and Challenges.}
Most AI systems remain reactive, responding after explicit queries or commands. Smart glasses continuously observe egocentric context and therefore offer a natural platform for proactive assistance, but this creates a tension between \emph{anticipatory intelligence and user autonomy}. \textbf{(1) Intent inference:} the system must infer latent needs from gaze, behavior, environment, and interaction history. \textbf{(2) Intervention timing:} assistance should appear only when useful and unlikely to interrupt ongoing activity. \textbf{(3) Assistance selection:} the system must decide what information or action to provide while avoiding irrelevant, redundant, or aggressive interventions.

\textbf{Outlook.}
\textbf{(1) Context-aware intent modeling} can combine multimodal signals and long-term user context to predict emerging needs before explicit requests. \textbf{(2) Adaptive intervention policies} should balance expected utility against interruption cost. \textbf{(3) User-controllable proactivity} should let users adjust, reject, or personalize proactive behaviors. Proactive smart glasses should provide timely assistance while preserving attention, control, and user agency.

\subsection{Embodied Foundation Models}
\label{sec:embodied-foundation-models}

\textbf{Status and Challenges.}
Existing foundation models perform strongly in language, vision, and multimodal reasoning, but are largely developed for static or short-context inputs rather than continuous embodied interaction. AI smart glasses expose a gap between \emph{multimodal understanding and embodied intelligence}: models must recognize what users see, track how the environment evolves, and relate observations to ongoing actions and goals. Three challenges are central. \textbf{(1) Egocentric world understanding:} models must reason about spatial relations, activities, affordances, and user intentions from egocentric observations. \textbf{(2) Temporal and task reasoning:} continuous interaction requires tracking state changes, action history, and task progress over long horizons. \textbf{(3) Efficient embodied inference:} these capabilities must operate with low latency and limited computation, memory, and energy.

\textbf{Outlook.}
\textbf{(1) Egocentric pretraining} on large-scale egocentric multimodal data can ground models in spatial, behavioral, and task-level knowledge. \textbf{(2) Perception--reasoning--action modeling} should jointly understand observations, infer user goals, and predict appropriate actions within continuous interaction loops. \textbf{(3) Efficient embodied models} should combine lightweight architectures, model compression, and edge-assisted inference to make real-time embodied intelligence practical on smart glasses. Embodied foundation models should make smart glasses capable of continuous understanding, reasoning, and assistance in real-world activity.

\section*{Conclusion}
\label{sec:conclusion}

AI smart glasses are becoming an important platform for wearable intelligence because they combine egocentric perception, hands-free interaction, continuous availability, and intelligent reasoning during daily activities. However, research on AI smart glasses remains dispersed across device design, model capabilities, interaction techniques, and domain-specific deployments, making a systematic overview necessary. To bridge this gap, this survey reviewed AI smart glasses through a system-level taxonomy spanning hardware foundation, wearable intelligence, interaction design, and application scenarios, providing researchers with a structured understanding of how wearable assistance is built, exposed to users, and adapted to real-world contexts. Since AI smart glasses are still in the early stages of development, we also discussed current limitations and future research directions to make such systems more deployable, trustworthy, personalized, and proactive. We hope this survey helps clarify the design space of AI smart glasses and supports future research on wearable intelligence in real-world settings.

\clearpage
\bibliographystyle{unsrtnat}
\bibliography{egbib}

\clearpage
\vfill

\end{document}